%% file: acl_latex.tex
\documentclass[11pt]{article}

\usepackage[final]{acl}

\usepackage{times}
\usepackage{latexsym}
\usepackage{amsmath}
\usepackage{amssymb}
\usepackage{bbm}
\usepackage{microtype}
\usepackage{graphicx}
\usepackage{amsfonts}
\usepackage{booktabs}
\usepackage{multirow}
\usepackage{hyperref}
\usepackage{cleveref}
\usepackage{pgfplotstable}
\usepackage{listings}
\usepackage{tcolorbox}
\usepackage{enumitem} 
\usepackage{subcaption} 
\usepackage{float}
\usepackage{cuted}
\usepackage{capt-of}
\usepackage{booktabs,tabularx,array}
\usepackage{xltabular}
\usepackage{pdfpages}
\usepackage{makecell}
\usepackage[section]{placeins}

\usepackage[T1]{fontenc}

\usepackage[utf8]{inputenc}

\usepackage{microtype}

\usepackage{inconsolata}

\usepackage{graphicx}

\newcolumntype{Y}{>{\raggedright\arraybackslash}X}

\title{Beyond Top Words: MonoTM for Topic Modeling with Interpretable Monosemantic Features}

\author{
Una Joh \and Bei Yu \\
School of Information Studies \\
Syracuse University \\
\texttt{\{sjoh01,byu\}@syr.edu}
}

\begin{document}
\maketitle
\begin{abstract}
Topic models summarize large text corpora, but top-ranked words often provide only a limited representation of topic semantics. Sparse autoencoders (SAEs) offer a way to move beyond word-level descriptors by extracting interpretable features from dense representations, yet how feature interpretability relates to topic-inference quality remains unclear. We introduce \textbf{MonoTM}, an interpretable topic modeling framework that decouples these roles. Across three benchmark corpora, we show that document--topic mixture estimation and semantic interpretation favor different SAE configurations and feature subsets. MonoTM estimates mixtures from the full SAE bag-of-features representation and, with them fixed, learns topic descriptors over a separate vocabulary of corpus-grounded semantic features. This design preserves global topic structure while representing topics with semantic units more meaningful than individual words, making them more useful for downstream corpus analysis.
\end{abstract}

\section{Introduction}

Text is a primary source of evidence across the social sciences and the humanities, from political speeches and news coverage to historical archives and literary corpora. The scale of these collections motivates computational text-as-data methods; topic modeling, especially Latent Dirichlet Allocation (LDA) and its extensions, has therefore become a popular tool for exploratory corpus mapping and thematic summarization \citep{blei_latent_2003}. Yet its role as scholarly evidence remains contested \cite{da_computational_2019, grimmer_text_2013, shadrova_topic_2021}. A recent review of topic modeling validation in computational social science finds that validation and reporting practices remain heterogeneous and show little convergence across studies \citep{bernhard-harrer_beyond_2025}.

One source of this issue is that most topic models define each topic as a probability distribution over words, and users typically interpret topics by their highest-probability words. However, assigning a concise, human-readable meaning or label to each inferred topic can be challenging in practice \citep{chang_reading_2009, mei_automatic_2007}. Although recent neural topic models incorporate pretrained language models or document embeddings to capture semantic information beyond raw word counts, these models still typically extract top-ranked words or phrases post hoc to describe each topic to users \citep{bianchi_pre-training_2021, dieng_topic_2020, grootendorst_bertopic_2022, kardos_s3_2025}.

This creates a gap between what topic models output—topics as distributions over words—and what many analysts want: semantic units that support interpretation at an intermediate level between reading individual documents and identifying coarse corpus-level themes. Beyond topic representation, words are rarely adequate semantic units for supporting meaningful downstream corpus analysis, such as analyzing relations between topics or documents based on word overlap.

Recent progress in mechanistic interpretability, particularly sparse autoencoders (SAEs) and autointerpretability methods, offers a way to extract human-readable semantic features from document embeddings produced by pretrained encoders. Motivated by our experimental findings, we propose \textbf{MonoTM}, an interpretable topic modeling framework that achieves competitive document--topic mixture estimation while producing validated, human-readable topic descriptors. \footnote{The official MonoTM repository is available at https://github.com/Una-J/MonoTM}

\section{Background \& Related Work}
\label{sec:related}

\subsection{Mechanistic Interpretability and Sparse Autoencoders}
\label{sec:mi_sae}

Mechanistic interpretability seeks to explain neural networks by identifying internal components
and how they compose into computations. A major obstacle is that neurons are often polysemantic, activating for multiple unrelated patterns \cite{olah_2020_zoom}. One influential account explains this via superposition, where models represent more features than they have dimensions by encoding features in an overcomplete set of directions,
enabled by sparsity in the underlying factors \citep{elhage_2022_toy}. This motivates moving from
neurons to learned feature directions.

Two scalable strategies support this shift. First, \citet{bills_language_2023} employ language models to produce and score natural-language explanations of internal directions based on activation patterns. Second, sparse dictionary learning methods, especially sparse
autoencoders (SAEs), learn sparse, overcomplete feature bases that often yield more monosemantic
units than neurons and can support meaningful interventions \citep{bricken2023monosemanticity, huben_sparse_2024}.
Recent work also applies SAEs to dense text embeddings, recovering interpretable semantic factors
useful for downstream control and analysis \citep{oneill_towards_2024}.

Building on these ideas, we investigate SAEs over document embeddings as a source of reusable semantic units for topic modeling, and adopt an Interpreter--Predictor style protocol to obtain validated feature labels for topic description.

\subsection{Topic Modeling}
\paragraph{Overview of topic models.}
Topic modeling began with probabilistic bag-of-words (BoW) generative models such as LDA \citep{blei_latent_2003},
which represents each document as a mixture of topics and each topic as a distribution over words.
Although LDA offers clear probabilistic semantics and interpretable topic descriptors, its BoW assumption ignores context and can produce frequency- and surface-form-sensitive topics. Neural topic models use inference networks for amortized variational inference instead of iterative per-document inference (ProdLDA; \citep{srivastava_autoencoding_2017}). Other neural parameterizations of document--topic mixtures include softmax and stick-breaking constructions \citep{miao_discovering_2017}.
 In parallel, embedding-aware models such as ETM represent topics in a word-embedding space to improve robustness to large vocabularies while maintaining interpretability \citep{dieng_topic_2020}.

More recent approaches incorporate pretrained transformers to inject contextual semantics. A representative direction is
contextualized topic modeling. 
CombinedTM uses contextual embeddings alongside BoW-style signals to improve topic quality \citep{bianchi_pre-training_2021},
and ZeroShotTM extends this idea with multilingual encoders to enable cross-lingual, zero-shot topic inference
\citep{bianchi_cross-lingual_2021}. 
Beyond contextualization, recent work focuses on robustness and efficiency. ECRTM introduces embedding clustering
regularization to discourage topic collapse and encourage more distinct topics \citep{wu_effective_2023}. FASTopic further
leverages pretrained transformers while modeling document--topic and topic--word relations in a unified framework
designed to be fast, stable, and transferable \citep{wu_fastopic_2024}. Finally, S3 offers a decomposition-based alternative that treats topics as independent semantic axes in embedding space, discovered via ICA rather than clustering or neural
decoders \citep{kardos_s3_2025}.

\paragraph{SAE-based topic modeling.}
Recent work connects sparse autoencoders and topic modeling by leveraging mechanistic-interpretability insights and treating SAE latents as reusable semantic units in embedding/activation space. \citet{girrbach_sparse_2025} view SAEs as continuous-space topic models by deriving the SAE objective as a MAP estimator under an LDA-like generative model. \citet{zheng_model_2025} propose Mechanistic Topic Models (MTMs), which featurize documents using a pretrained SAE over LLM activations and fit topic models over SAE feature counts. While their mLDA is closest to our LDA-with-bag-of-features approach, MTMs rely on pretrained SAEs and external feature descriptions.

Unlike these prior works, our goal is not simply to replace words with SAE latents as the vocabulary for topic modeling. We show that the features most useful for document--topic mixture estimation are not necessarily the same features whose labels can be reliably interpreted. This empirical mismatch motivates MonoTM, which decouples the statistical and interpretive roles of SAE features. Although this design is more costly than reusing pretrained SAE vocabularies, it targets settings such as social science and digital humanities, where corpus-specific interpretability, traceability, and validation are often more important than fast topic discovery.

\section{Research Questions}
\label{sec:rq}

Our starting hypothesis is that SAE features can play two distinct roles in topic modeling. 
If a document embedding can be represented as a sparse, nonnegative linear combination of SAE features, and if those features are closer to monosemantic than the original embedding dimensions, then SAE features may serve both as human-readable semantic units and as useful units for estimating document--topic mixtures. These two roles are related but not identical: features that are easy to label may not be the same features that best support mixture estimation. We therefore first study these roles separately, and then ask how they can be combined in a single interpretable topic model:

\begin{enumerate}
  \item Under what SAE configurations do corpus-trained features become reliably interpretable semantic units?
  \item Under what SAE configurations do SAE features provide effective units for document–topic mixture estimation?
  \item How can we use SAE features to build a topic model that produces interpretable semantic topic descriptors aligned with strong document–topic mixture estimates?  
\end{enumerate}

\section{Experimental Setup}
\label{sec:setup}

\paragraph{Datasets.}
To evaluate topic models with minimal subjectivity, we compare inferred topics against the gold category annotations provided by standard benchmarks (dataset statistics in Table~\ref{tab:dataset_stats_pgf}). We use \textbf{(1) 20 Newsgroups} (user-generated Usenet newsgroup posts; scikit-learn version) \citep{lang1995newsweeder}, \textbf{(2) Web of Science} (WOS-46985) using abstracts as input \citep{kowsari2017HDLTex}, and \textbf{(3) Reuters} (Reuters-21578) with the ModApte split \citep{lewis1997reuters} accessed via Hugging Face Datasets \citep{lhoest_datasets_2021}. Reuters is a multi-label and highly imbalanced collection of Reuters newswire stories. For Reuters, we keep labels appearing at least 30 times, yielding 47 topics.

\paragraph{Document embeddings and SAE training.}
We map each document $d$ to a dense embedding $x_d \in \mathbb{R}^{4096}$ using NVIDIA's
\texttt{llama-embed-nemotron-8b} model~\cite{babakhin_2025_llamaembednemotron8buniversaltextembedding}.
We use the model's full context window of 32{,}768 tokens~\cite{nvidia_embed_modelcard}; longer documents are truncated to their first 32{,}768 tokens. This affects only 18 documents in 20 Newsgroups.

For each corpus, we train top-$K$ sparse autoencoders on dimension-wise standardized document embeddings. Each SAE learns a dictionary with $mN$ latent features, where $m=4096$ is the embedding dimension and $N$ is the expansion factor. Thus, $N$ controls the size of the learned feature dictionary, while $K$ controls the maximum number of active features per document. Given a document embedding, the encoder produces nonnegative latent activations and retains only the largest $K$, yielding a sparse document--feature representation. The decoder reconstructs the standardized embedding from these active features.

The two SAE hyperparameters therefore have distinct roles: $N$ controls dictionary capacity, while $K$ controls per-document activity. We sweep 11 values of $N$ from $1/64$ to $5$ and eight values of $K$ from $4$ to $512$, omitting configurations with $mN<K$. Full architectural, optimization, checkpoint-selection, and grid details are given in Appendix~\ref{app:sae-training}.

\paragraph{Interpreter: feature label generation.}
Following and adapting the autointerpretability procedure of \citet{oneill_towards_2024}, we pass the learned SAE features (decoder columns) to an autointerpretability module that proposes and later validates natural-language labels.

For each feature $f\in\{1,\dots,mN\}$, we construct an Interpreter prompt using three sets of examples: \textbf{Max-activating examples}, the 10 documents with the highest activations $h_{d,f}$; \textbf{Typical-activating examples}, 10 documents sampled from the middle quantiles of the nonzero activation distribution for $f$; and \textbf{Zero-activating examples}, 10 documents where feature $f$ is inactive ($h_{d,f}=0$), sampled as negatives.

Given these examples, an \emph{Interpreter LLM} outputs a short description $\ell_f$ (e.g., 4--10 words) intended to capture the single most prominent concept that is present in the activating texts but absent from the non-activating texts (see the Interpreter prompt in Figure~\ref{fig:full_prompts}).

\paragraph{Predictor: feature label validation.}
Labels produced by an Interpreter LLM can be plausible but unvalidated. For example, labels may describe a frequent corpus theme that is not specific to the feature. To test whether a label actually predicts feature behavior, we validate each $\ell_f$ using a separate Predictor LLM \citep{bills_language_2023,oneill_towards_2024}.

For each feature $f$, we construct a balanced evaluation set $\mathcal{S}_f=\mathcal{S}_f^{+}\cup\mathcal{S}_f^{-}$ by sampling 30 \emph{positive} documents and 30 \emph{negative} documents from the corpus. Positives satisfy $h_{d,f}>0$ (feature $f$ is active for document $d$), while negatives satisfy $h_{d,f}=0$ (feature $f$ is inactive).
Let $y_{d,f}=\mathbf{1}[h_{d,f}>0]$ denote the ground-truth activation indicator.

We run the Predictor as an independent per-document binary classification query. For each $d\in\mathcal{S}_f$, the Predictor LLM receives only the description $\ell_f$ and the raw document text, and outputs a binary prediction $\hat{y}_{d,f}\in\{0,1\}$ indicating whether feature $f$ would activate (see the Predictor prompt in Figure~\ref{fig:full_prompts}). We aggregate these document-level predictions into a feature-level interpretability score using the F1 score between $\{\hat{y}_{d,f}\}_{d\in\mathcal{S}_f}$ and $\{y_{d,f}\}_{d\in\mathcal{S}_f}$, and denote this score by $\mathrm{IS}(f)$.

\paragraph{Autointerpretability sweep and LLM setup.}
Running the Interpreter--Predictor protocol requires many LLM inference calls, so we apply it only to a subset of SAE configurations we expect to be most informative due to cost constraints:
$N\in\{0.5,1,2,3,4,5\}$ and $K\in\{4,8,16,32\}$. Details of the LLMs used are in Appendix~\ref{sec:appendix_llm_detail}.

\section{Results}
\label{sec:results}

\subsection{RQ1: When Do SAE Features Become Interpretable Semantic Units?}
\label{sec:rq1_results}

RQ1 asks whether corpus-trained SAE features can serve as reliable semantic units, and how this depends on SAE capacity and sparsity. We operationalize interpretability using the Interpreter--Predictor protocol described in Section~\ref{sec:setup}: a feature is treated as interpretable when its natural-language label predicts held-out feature activations with an interpretability score above a validation threshold.

\paragraph{Empirical results.}
SAE features become interpretable semantic units most consistently under low-to-moderate per-document activity levels. Across all three datasets, configurations with smaller $K$ have the highest fraction of activated features that pass validation (Tables~\ref{tab:rq1-20news}, \ref{tab:rq1-wos}, and~\ref{tab:rq1-reuters}).

Increasing $K$ changes this behavior. At permissive validation thresholds, larger $K$ generally increases the absolute number of validated features, because more active features are exposed for labeling and validation. However, when averaged across the $N$ values we evaluate, the fraction of activated features that are validated decreases monotonically with $K$ on all datasets (Tables~\ref{tab:rq1-20news}, \ref{tab:rq1-wos}, and~\ref{tab:rq1-reuters}).

The effect of dictionary capacity $N$ is best understood conditional on the activity level $K$. On 20 Newsgroups and Reuters, increasing $N$ at low $K$ often reduces the number of activated features that surpass the activation threshold, while Web of Science is an exception: activated and validated counts increase with $N$ even at low $K$ (Tables~\ref{tab:rq1-20news}, \ref{tab:rq1-wos}, and~\ref{tab:rq1-reuters}). This suggests that the value of increasing dictionary capacity depends on whether the corpus and activity level provide enough support for the larger feature space. At higher $K$, larger dictionaries are more often useful, as allowing documents to activate more features enables higher-capacity SAEs to expose a larger pool of validated semantic units.

Additionally, we report a secondary diagnostic analysis of feature granularity in Appendix~\ref{sec:granularity_probe_details}. This analysis suggests that increasing capacity and activity may shift the validated feature space toward finer semantic resolution.

\paragraph{Choosing $(N, K)$ in practice.}
These results suggest that SAE configuration should be treated as a practical modeling choice. If the user wants a conservative, easy-to-audit feature vocabulary, the best starting point is a low- or moderate-activity SAE, such as $K=4$ or $K=8$, combined with a moderately large dictionary. This setting yields fewer features, but a larger share of them can be validated. If the user instead wants broader coverage or more fine-grained descriptors, it is reasonable to move toward larger $K$ and larger $N$, but only with the expectation that post-hoc validation will discard a larger fraction of the activated feature space.

A useful practical procedure is therefore to begin with a high-precision anchor configuration and then expand only if the validated feature inventory is too coarse or too small. For corpora with many stable domain-specific distinctions, such as Web of Science, a moderate configuration such as $K=8$ with a larger dictionary is a natural starting point. For noisier or more heterogeneous corpora such as 20 Newsgroups and Reuters, low-$K$ configurations provide a cleaner initial semantic inventory.

\subsection{RQ2: When Do SAE Features Support Document--Topic Mixture Estimation?}
\label{sec:rq2_results}

Parallel to RQ1's focus on interpretability, RQ2 asks whether corpus-trained SAE features can serve as useful statistical units for estimating document--topic mixtures.

We test this hypothesis by replacing LDA's word vocabulary with SAE features. For each document, we treat active SAE features as pseudo-tokens and their activation magnitudes as token weights, producing a bag-of-features (BoF) representation. We deliberately choose LDA because it provides a transparent mixed-membership likelihood once document embeddings are converted into bags of SAE features. Neural topic models could also be applied to the same BoF inputs, but their neural parameterization would introduce additional representational capacity, making it harder to isolate whether the SAE feature vocabulary itself provides useful statistical units for topic inference.

RQ2 evaluates whether SAE features are useful for mixture estimation, regardless of their interpretability scores.

\paragraph{BoF+LDA setup.}
For each dataset, we train a top-$K$ SAE and construct a sparse document--feature matrix
$X\in\mathbb{R}_{\ge 0}^{D\times mN}$ where $X_{d,j}$ is the activation magnitude of feature $j$
in document $d$.
We then fit LDA using scikit-learn's \texttt{LatentDirichletAllocation} with variational EM for 200 iterations, and set the number of topics to the number of gold categories
(20 Newsgroups: $T{=}20$, Web of Science: $T{=}7$, Reuters: $T{=}47$).

\paragraph{Baselines.} We compare against (i) classical BoW-LDA \citep{blei_latent_2003} and (ii) other state-of-the-art neural topic models that output document--topic mixtures: CombinedTM \citep{bianchi_pre-training_2021}, ZeroShotTM \citep{bianchi_cross-lingual_2021}, ECRTM \citep{wu_effective_2023}, FASTopic \citep{wu_fastopic_2024}, and S$^3$ \citep{kardos_s3_2025}. We do not include BERTopic because, following the taxonomy of \citet{wu_survey_2024}, it is a clustering-based topic discovery method rather than a model for document--topic mixture estimation. Including it in the topic--label alignment benchmark would therefore conflate mixture-estimation quality with embedding-cluster separability. 

For each baseline, we apply the same topic--label alignment procedure as for BoW+LDA. To ensure a fair comparison, all baselines that use document embeddings---CombinedTM, ZeroShotTM, FASTopic, and S$^3$---are run with the same \texttt{llama-embed-nemotron-8b} document embeddings used by MonoTM.

\paragraph{Evaluation.}
We do not use standard word-based topic coherence or topic diversity as evaluation metrics because doing so would defeat the purpose of MonoTM. Metrics such as PMI, NPMI, $C_V$, and topic diversity were designed for topic representations expressed as ranked lists of keywords: they measure whether top words co-occur in a reference or training corpus, or whether top-word lists are lexically redundant across topics \citep{newman_external_2009,mimno_optimizing_2011,lau_machine_2014,roder_exploring_2015,dieng_topic_2020}. These metrics would require projecting our descriptors back into keywords that represent each topic, thereby reintroducing the very representation that our method is designed to replace.

We instead evaluate document--topic mixture quality by aligning inferred topics with gold benchmark labels and reporting Micro-F1 and Macro-F1. Full evaluation details, including the single-label and multi-label alignment procedures, are given in Appendix~\ref{sec:eval_details}. For all models and datasets, we run three trials with different random seeds and report mean scores in Table~\ref{tab:classification_results}; full SAE-grid heatmaps are reported in Appendix~\ref{sec:rq2_full_heatmaps}.

\subsubsection{Effective mixture estimation requires balanced SAE scaling}
\label{sec:obsB_theta_scale}

BoF+LDA performance depends systematically on the SAE hyperparameters. The full Micro-F1 and Macro-F1 heatmaps are reported in Appendix~\ref{sec:rq2_full_heatmaps}. Across datasets, mixture quality is weakest in two regimes: very wide dictionaries paired with very small $K$, and boundary cases where $mN=K$. These results suggest that SAE features are most useful for topic inference when the representation maintains an explicit sparsity bottleneck while still allowing enough active features per document.

The best-performing configurations instead lie in a balanced scaling band where dictionary capacity and per-document activity increase together. We therefore summarize BoF+LDA in Table~\ref{tab:classification_results} by averaging over a fixed robust subset of this region, using $N\in\{0.25,0.5,1,2\}$ and $K\in\{128,256,512\}$. This gives a less brittle comparison than selecting the single best hyperparameter setting for each dataset.

\begin{table*}[t]
\centering
\small
\begin{tabular}{lcccccc}
\toprule
\multirow{2}{*}{\textbf{Model}} & \multicolumn{2}{c}{\textbf{20 Newsgroups}} & \multicolumn{2}{c}{\textbf{Web of Science}} & \multicolumn{2}{c}{\textbf{Reuters}} \\
\cmidrule(lr){2-3} \cmidrule(lr){4-5} \cmidrule(lr){6-7}
 & Micro F1 & Macro F1 & Micro F1 & Macro F1 & Micro F1 & Macro F1 \\
\midrule
BoW-LDA & 0.3673 & 0.3515 & 0.5147 & 0.4702 & \textbf{0.3446} & 0.2021 \\
CombinedTM & 0.4929 & 0.4513 & 0.5514 & 0.5201 & 0.2398 & 0.1728 \\
ZeroShotTM & 0.3704 & 0.3398 & 0.4904 & 0.4670 & 0.2051 & 0.1467 \\
ECRTM & 0.5774 & 0.5368 & 0.5176 & 0.4916 & 0.2410 & 0.1557 \\
FASTopic & 0.5492 & 0.4912 & 0.5032 & 0.4583 & 0.2535 & 0.1582 \\
S$^{3}$ & 0.4578 & 0.4220 & 0.4233 & 0.3942 & 0.1513 & 0.1100 \\
\midrule
BoF+LDA (Ours) & \textbf{0.6770} & \textbf{0.6489} & \textbf{0.5753} & \textbf{0.5446} & 0.3188 & \textbf{0.2089} \\
\bottomrule
\end{tabular}
\caption{Micro F1 and Macro F1 scores on 20 Newsgroups, Web of Science, and Reuters datasets. The best scores are in bold.}
\label{tab:classification_results}
\end{table*}

\subsubsection{Mixture quality and interpretability do not fully coincide}
\label{sec:filter}

The balanced-scaling pattern identified above should not be read as an interpretability result. Although Figure~\ref{fig:rq2heatmaps} identifies a robust region for document--topic mixture estimation, the same region is not necessarily the one that maximizes the number of interpretable SAE features. This can be seen by comparing the mixture-quality heatmaps in Figure~\ref{fig:rq2heatmaps} with the interpretability counts in Tables~\ref{tab:rq1-20news}, \ref{tab:rq1-wos}, and~\ref{tab:rq1-reuters}. For example, in 20 Newsgroups, holding $N=0.5$ fixed and increasing $K$ from $8$ to $32$ reduces the fraction of dictionary features with $\mathrm{IS}\geq 0.80$ from $721/(4096\cdot0.5)=35.2\%$ to $282/(4096\cdot0.5)=13.8\%$, while Micro/Macro F1 increases from 0.56/0.52 to 0.66/0.63. Thus, neither the absolute number nor the dictionary-level fraction of validated features fully predicts document--topic mixture quality. These results suggest a tension between choosing SAE configurations for feature-level interpretability and choosing them for document--topic mixture quality.

We test this tension more directly by asking whether restricting BoF+LDA to highly interpretable SAE features improves document--topic mixture estimation. Specifically, we filter features according to their interpretability score, retain only those with $\mathrm{IS}(f)\ge\tau$, and then re-fit LDA on the resulting filtered BoF representation. Figures~\ref{fig:20ngf1fiter}--\ref{fig:reutersf1fiter} show that this filtering generally hurts document--topic mixture quality, and that the degradation becomes more severe as $\tau$ increases. These results indicate that failing strict interpretability validation does not imply that an SAE feature is uninformative for mixture estimation.

\subsubsection{Bag-of-features yields strong document--topic mixture estimates}
\label{sec:rq2_1}

Table~\ref{tab:classification_results} shows that BoF+LDA is competitive across all three benchmarks, achieving the strongest results on both single-label datasets and the best Macro-F1 on Reuters. These gains support the core intuition that SAE latents provide a sparse, semantically structured pseudo-token vocabulary, allowing LDA to pool co-occurrence evidence more effectively than when using raw words.

Results on Reuters are nuanced. Since it is multi-label and highly imbalanced, Micro-F1 is dominated by frequent classes while Macro-F1 is sensitive to performance on rare labels. Here, BoW-LDA attains the highest Micro-F1, but BoF+LDA achieves the best Macro-F1 overall. This pattern is consistent with BoF features providing more discriminative evidence for minority topics that have limited word overlap with majority classes.

\subsection{RQ3: Interpretable Topic Modeling (MonoTM)}
\label{sec:MonoTM}

RQ3 asks how the two roles of SAE features identified above can be combined into a single interpretable topic model. The findings from RQ1 and RQ2 suggest that a single SAE representation should not be forced to serve both feature-level interpretability and document--topic mixture estimation. MonoTM implements this idea by separating document--topic mixture estimation from topic interpretation.

The remainder of this section presents MonoTM in three steps. Section~\ref{sec:MonoTM-algorithm} describes the MonoTM algorithm. Section~\ref{sec:excluded-feature-audit} audits whether the validated descriptor vocabulary omits important topic semantics from lower-validation features. Section~\ref{sec:descriptor-comparison} gives a compact descriptor comparison against word-based and free-form LLM descriptor baselines.

\subsubsection{The MonoTM Algorithm}
\label{sec:MonoTM-algorithm}

MonoTM separates the statistical and interpretive roles of SAE features. Let \(D\) be the number of documents, \(T\) the number of topics, and \(V\) the number of validated interpretable features used for topic description. MonoTM estimates document--topic mixtures from a full SAE bag-of-features representation, then estimates a topic--feature distribution over validated interpretable features with the mixtures held fixed.

\paragraph{Stage 1: estimating document--topic mixtures from all SAE features.}
For each document, the mixture SAE produces a sparse nonnegative activation vector over all SAE features. We construct a full document--feature matrix \(X^{\mathrm{mix}}\) from these activations and fit LDA with \(T\) topics, yielding document--topic mixtures
\[
\Theta \in \mathbb{R}^{D \times T}_{\geq 0},
\qquad
\theta_d \in \Delta^T .
\]
This stage uses all active SAE features, rather than only validated interpretable features, because Section~\ref{sec:filter} shows that filtering to interpretable features degrades mixture quality.

\paragraph{Stage 2: constructing an interpretable document--feature matrix.}
For topic interpretation, we use a validated feature set \(\mathcal{F}_{\tau}\). A feature is retained if it activates in at least 30 documents, has a nonempty label, and satisfies \(IS(f) \geq \tau\), with \(\tau=0.8\) in our experiments. For each retained feature \(f\), we construct an interpretable document--feature matrix \(C \in \mathbb{R}^{D \times V}_{\geq 0}\) by weighting its activation by its interpretability score:
\[
c_{d,f} = \max(\tilde{h}_{d,f},0)\cdot IS(f),
\qquad f \in \mathcal{F}_{\tau}.
\]

\paragraph{Stage 3: estimating topic--feature distributions with fixed mixtures.}
Given fixed document--topic mixtures \(\Theta\) and interpretable feature matrix \(C\), MonoTM estimates a topic--feature distribution
\[
B \in \mathbb{R}^{T \times V}_{\geq 0},
\qquad
\beta_t \in \Delta^V ,
\]
by maximizing the fixed-mixture likelihood
\[
\mathcal{L}(B)
=
\sum_{d=1}^{D}
\sum_{f=1}^{V}
c_{d,f}
\log
\left(
\sum_{t=1}^{T}
\theta_{d,t}\beta_{t,f}
\right).
\]
We optimize this objective with a Dirichlet-smoothed EM procedure; full update equations and implementation details are given in Appendix~\ref{sec:MonoTM-details}.

After estimating \(B\), each topic \(t\) is represented by the highest-probability validated feature labels under \(\beta_t\). Appendix~\ref{sec:sample-MonoTM-output} provides sample MonoTM topic--feature representations produced by this final descriptor layer.

\subsubsection{Audit: do excluded features hide important topic semantics?}
\label{sec:excluded-feature-audit}

To test whether the validated descriptor vocabulary omits important topic semantics, we audit the excluded features most likely to matter for topic interpretation. For all three datasets, we estimate \(\Theta\) from the full bag-of-features representation of an SAE with \(N=0.5\) and \(K=256\). For interpretation, we use the $N=2, K=16$ SAE. The final MonoTM descriptor vocabulary contains active, validated features with $\mathrm{IS}(f)\ge .8$. For the audit, we estimate an auxiliary all-feature topic--feature distribution $B^{\mathrm{all}}$ over all active interpretation-SAE features, using raw feature activations instead of weighting by $\mathrm{IS}(f)$. This avoids mechanically suppressing the lower-validation features that the audit is designed to inspect.

We then rank all active interpretation features by their mass under $B^{\mathrm{all}}$ for each topic, and focus on lower-validation features with $\mathrm{IS}(f)<.8$ that would have ranked among the top 20 features for at least one topic. These features are the strongest possible challenge to the validated descriptor vocabulary: they are highly topic-associated, but absent from the final MonoTM representation. For each such feature, assigned to the topic where it obtains its best rank, we search for a validated same-topic neighbor among the top-50 validated features of that topic under $B^{\mathrm{all}}$. Appendix~\ref{sec:excluded-feature-audit-details} gives the full construction, row alignment, EM objective, filtering rules, and formal definitions.

\begin{figure}[t]
\centering
\includegraphics[width=\linewidth]{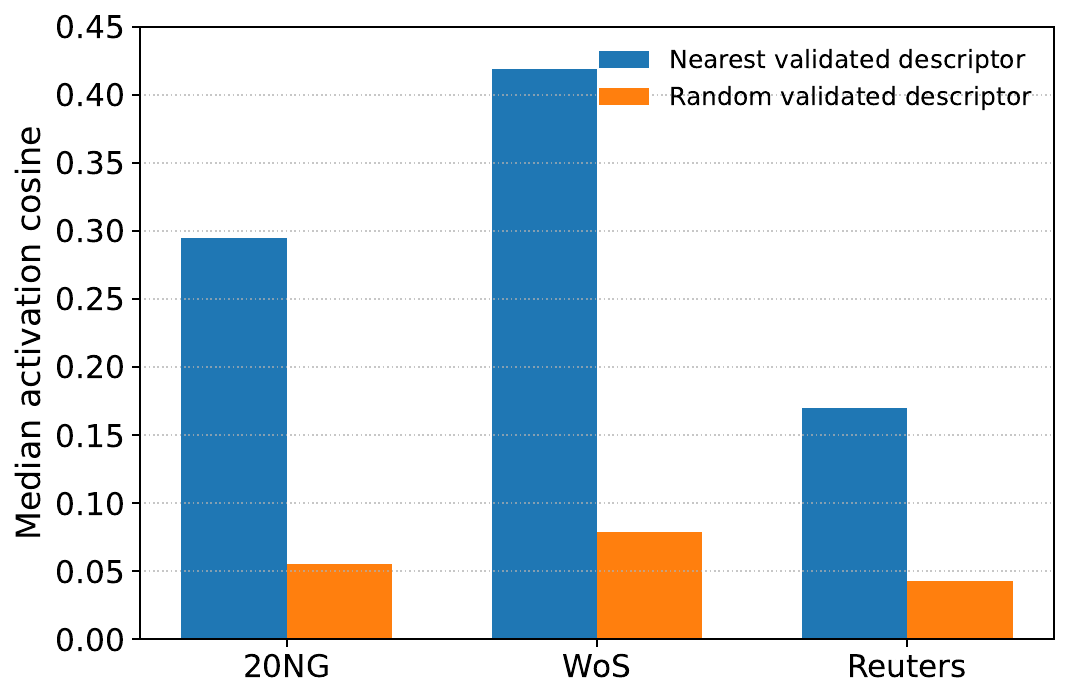}
\caption{Coverage of lower-validation high-association features by validated same-topic descriptors. Bars show the median cosine similarity between feature activation profiles across documents.}
\label{fig:excluded-feature-coverage}
\end{figure}

Coverage is measured in document-activation space. For each excluded feature, we compute cosine similarity between its raw document-level interpretation-SAE activation vector and the corresponding vectors for validated same-topic descriptors. We compare the nearest validated same-topic descriptor to a random validated descriptor sampled from the same topic-specific candidate pool. Figure~\ref{fig:excluded-feature-coverage} shows that nearest validated descriptors are much closer than random validated descriptors, especially for 20 Newsgroups and Web of Science.

The Reuters results are less straightforward, so we further characterize the excluded high-association features using heuristic label-type flags based on keywords in the generated feature labels. (See Appendix~\ref{sec:excluded-feature-audit-details} for the keyword rules.) These flags are used to summarize broad tendencies in the excluded feature set, not as a validation metric.

Table~\ref{tab:excluded-feature-label-types} shows a clear dataset difference. In 20 Newsgroups and Web of Science, excluded high-association features are mostly semantic or broad-discourse labels. In Reuters, by contrast, 70.7\% are flagged as format/register/numeric cues, suggesting that many of these features capture financial-news register, reporting format, or numeric conventions rather than missing core topic semantics. Concrete examples of both patterns are provided in Appendix~\ref{sec:excluded-feature-audit-details}.

\begin{table}[t]
\centering
\small
\begin{tabular}{lrrr}
\toprule
Dataset & Semantic & \makecell{Broad\\discourse} & \makecell{Format/register/\\numeric} \\
\midrule
20NG & 58.1\% & 36.4\% & 5.4\% \\
WoS & 88.5\% & 7.7\% & 3.8\% \\
Reuters & 6.3\% & 23.1\% & 70.7\% \\
\bottomrule
\end{tabular}
\caption{Heuristic label-type split for lower-validation high-association features. Percentages are computed within each dataset.}

\label{tab:excluded-feature-label-types}
\end{table}

Overall, while this audit does not prove that no excluded feature ever captures a meaningful omitted semantic dimension, it alleviates the concern that MonoTM arbitrarily omits a large class of independent topic-defining features.

\subsubsection{Descriptor Comparison}
\label{sec:descriptor-comparison}

Finally, we compare descriptor mechanisms under the same fixed document--topic mixtures \(\Theta\). This comparison is intended to isolate the descriptor layer by asking how different mechanisms describe the same inferred topics. We compare MonoTM against three descriptor baselines: \(\theta\)-weighted TF-IDF terms, BERTopic-style c-TF-IDF terms~\citep{grootendorst_bertopic_2022}, and free-form LLM-generated descriptors. We include the LLM baseline because recent work increasingly treats LLMs as topic extractors~\citep{lam_concept_2024,mu_large_2024,pham_topicgpt_2024}. Full descriptor-generation details, prompts, and topic-name compression settings are provided in Appendix~\ref{sec:descriptor-baseline-details}. Full topic-name comparison tables are provided in Appendix~\ref{app:rq3-topic-name-tables}, and representative descriptor lists before topic-name compression are provided in Appendix~\ref{app:descriptor-examples}.

The comparison suggests two qualitative patterns. First, word-based descriptors often recover salient lexical anchors but remain surface-level. For example, in 20 Newsgroups, for the topic aligned with \texttt{talk.politics.misc}, the word-based baselines produce the topic name ``Gun rights,'' whereas MonoTM yields ``American culture wars,'' reflecting a broader mixture of gun-rights, sexuality, morality, and civil-liberties discussions (Table~\ref{tab:rq3-topic-names-20ng-app}). Second, free-form LLM descriptors can be fluent but overly local to the sampled topic-associated documents. In Reuters, for example, the LLM baseline names the topic aligned with \texttt{crude} as ``Ecuador oil crisis,'' whereas MonoTM yields the broader ``OPEC oil market'' (Table~\ref{tab:rq3-topic-names-reuters-app}). These examples suggest that MonoTM descriptors offer a useful intermediate level of semantic abstraction: they bridge high-level topic labels and raw documents more effectively than word lists, while remaining more directly tied to the representations used for topic inference than ad hoc LLM descriptors.

Beyond topic naming, MonoTM also supports downstream analyses of relations among topics. Appendix~\ref{app:semantic-feature-affordance} illustrates this use case by comparing topic relations in the validated-feature space.

\section{Conclusion}

We introduced \textbf{MonoTM}, an interpretable topic modeling framework built on sparse autoencoders trained over document embeddings. Across three research questions, our results show that SAE features are useful for topic modeling in multiple but distinct ways. First, corpus-trained SAEs can produce relatively monosemantic units, but the availability of reliably interpretable features depends strongly on the choice of SAE configuration. Second, SAE activations can also serve as effective statistical units for estimating document--topic mixtures. However, the SAE configurations suited to these two roles do not necessarily coincide. Addressing RQ3, MonoTM resolves this mismatch by estimating document--topic mixtures from the full SAE representation and then learning topic--feature distributions over validated features with the mixtures fixed. Our analyses further show that excluded high-association features are typically covered by validated same-topic descriptors or reflect auxiliary non-core cues.

\section{Limitations}
\label{sec:limit}

\paragraph{Remaining uncertainty about excluded features.}
Section~\ref{sec:excluded-feature-audit} audits features from the interpretation SAE that fall below the validation threshold and finds that genuinely hard-to-validate features account for only a small share of high-association topic-feature mass. However, this audit does not prove that all excluded features are unimportant or fully understood. Some low-validation features may encode corpus-specific regularities, pragmatic cues, formatting patterns, or culturally specific concepts that are difficult to summarize with short semantic labels. Moreover, our validation scores depend on the Interpreter and Predictor LLMs, so systematic blind spots in those models could still affect which features are included in the final descriptor set. For high-impact applications, MonoTM should therefore be paired with additional coverage audits and, when appropriate, human expert review.

\paragraph{Sensitivity to the Interpreter and Predictor LLMs.}
Our interpretability scores depend on the capabilities and failure modes of the specific LLMs used as Interpreter and Predictor, as well as prompt details and decoding settings. Different LLM backends may vary in (i) their ability to abstract from examples into stable hypotheses, (ii) calibration on the binary prediction task, and (iii) robustness to domain-specific language. As a result, both the number of features that validate above a threshold and the apparent granularity of validated features could change with the choice of model(s). While we partially mitigate this by separating generation (Interpreter) from validation (Predictor), our validation remains an LLM-mediated measurement rather than a ground-truth guarantee of monosemanticity.

\paragraph{Cost and practicality of LLM-heavy autointerpretability.}
MonoTM’s feature labeling relies on many LLM calls, which is computationally expensive and can be slow in practice. This cost is a real barrier for iterative modeling workflows. That said, MonoTM is aimed at settings where researchers are willing to spend more compute to obtain a single, carefully audited interpretive lens on an important corpus (e.g., social science and digital humanities analyses where interpretability and traceability are first-order goals). Still, improving efficiency is crucial. Promising directions include cheaper feature screening models for ``likely interpretable'' features and distilling the Predictor into a small classifier. Additionally, continuing progress in SAE feature interpretation and tooling \citep[e.g.,][]{lieberum_gemma_2024} suggests that MonoTM could become cheaper to apply while preserving its core advantages.

\paragraph{Systematic biases can distort the interpretable topic representation.}
MonoTM estimates document--topic mixtures using the full SAE vocabulary but represents topics using only the subset of features that pass LLM-based validation. If the Interpreter or Predictor has systematic blind spots---for example, consistently under-interpreting certain registers, dialects, domains, or culturally specific concepts---then those features may be disproportionately excluded from the interpretable feature set. In that case, the topic descriptors produced by MonoTM could omit important aspects of a topic even if those aspects strongly influence the inferred mixtures. This limitation is especially salient if the LLMs’ biases correlate with sensitive attributes or with particular styles of expression in the corpus. Mitigations include auditing interpretability coverage across document subpopulations, validating with multiple LLM backends, and incorporating human expert review for high-impact analyses.

\bibliography{custom}

\appendix

\section{Dataset Details}
\label{sec:datasets}
See Table~\ref{tab:dataset_stats_pgf} for summary statistics of the datasets.

\pgfplotstableread[col sep=comma]{dataset_stats_latex.csv}\datasetStats
\begin{table*}[t]
\centering
\small
\setlength{\tabcolsep}{3pt} 
\pgfplotstabletypeset[
  col sep=comma,
  string type,
  column type=c,
  columns={Dataset, NumDatapoints, NumCategories, MultiLabel, NumNoLabel, MeanChars, SDChars, MedianChars, MinChars, MaxChars},
  columns/Dataset/.style={string type, column type=l, column name={Dataset}},
  columns/NumDatapoints/.style={int detect, column type=r, column name={Points}},
  columns/NumCategories/.style={int detect, column type=r, column name={Cats}},
  columns/MultiLabel/.style={string type, column type=c, column name={Multi}},
  columns/NumNoLabel/.style={int detect, column type=r, column name={No Label}},
  columns/MeanChars/.style={fixed, precision=2, column type=r, column name={Mean Chars}},
  columns/SDChars/.style={fixed, precision=2, column type=r, column name={SD Chars}},
  columns/MedianChars/.style={int detect, column type=r, column name={Median}},
  columns/MinChars/.style={int detect, column type=r, column name={Min}},
  columns/MaxChars/.style={int detect, column type=r, column name={Max}},
  every head row/.style={
    before row=\toprule,
    after row=\midrule
  },
  every last row/.style={after row=\bottomrule},
]\datasetStats
\caption{Summary statistics of the  datasets used. The abbreviations in the header are defined as follows: Points (Total number of data points), Cats (Number of categories), and Multi (Multi-label classification, where 'O' denotes multi-label and 'X' denotes single-label). No Label indicates the number of samples without a label. The character count statistics (Mean, SD, Median, Min, Max) describe the distribution of document lengths.}
\label{tab:dataset_stats_pgf}
\end{table*}

\section{SAE Architecture and Training Details}
\label{app:sae-training}

For each document $d \in \mathcal{D}=\{1,\dots,M\}$, let $x_d \in \mathbb{R}^m$ denote its document embedding and let $\tilde{x}_d \in \mathbb{R}^m$ denote the dimension-wise standardized embedding, using the corpus-level mean and standard deviation. In our experiments, $m=4096$.

We use a single-layer top-$K$ sparse autoencoder with hidden dimensionality $mN$, where $N$ is the expansion factor. Depending on $N$, the dictionary may be undercomplete, complete, or overcomplete relative to the embedding dimension. The decoder is a matrix $D \in \mathbb{R}^{m \times (mN)}$, and the encoder uses tied weights $D^\top$:
\begin{align}
z &= D^\top \tilde{x} + b, \\
a &= \mathrm{ReLU}(z).
\end{align}
Sparsity is enforced with a hard top-$K$ operator,
\begin{equation}
h = \mathrm{TopK}(a, K),
\end{equation}
which yields a nonnegative latent activation vector $h \in \mathbb{R}^{mN}_{\ge 0}$ with at most $K$ active coordinates.

We additionally introduce a learnable per-feature gain vector $g \in \mathbb{R}^{mN}_{>0}$ to modulate feature magnitudes independently of decoder direction norms. We parameterize $g$ in log-space to enforce positivity and define
\begin{equation}
\tilde{h} = h \odot g.
\end{equation}
The reconstruction of the standardized embedding is
\begin{equation}
\hat{x} = D\tilde{h}.
\end{equation}
After each update, decoder columns are normalized to unit norm to avoid arbitrary rescaling between decoder weights and latent activations.

We train each SAE with Adam using learning rate $10^{-4}$ and batch size 1024 for up to 300k steps. We select the checkpoint with the lowest normalized reconstruction error on a fixed monitoring split containing 10\% of documents. The training objective is
\begin{equation}
\mathcal{L} =
\frac{\|\tilde{x}-\hat{x}\|_2^2}{\|\tilde{x}-\bar{x}\|_2^2+\varepsilon},
\end{equation}
where $\bar{x}$ is the batch mean.

We train a grid of top-$K$ SAEs over $(N,K)$:
\[
\begin{aligned}
N &\in \{0.015625, 0.03125, 0.0625, 0.125,\\
  &\phantom{\in \{} 0.25, 0.5, 1, 2, 3, 4, 5\},\\
K &\in \{4, 8, 16, 32, 64, 128, 256, 512\}.
\end{aligned}
\]
This gives up to $11 \times 8 = 88$ configurations. We omit configurations with $mN<K$, since they cannot activate $K$ distinct features. When $mN=K$, the top-$K$ operator retains all units, so the model reduces to a dense tied-weight autoencoder; we keep this boundary case as a diagnostic comparison.

\section{Prompts for the Interpreter–Predictor Module}
See Figure~\ref{fig:full_prompts} for the Interpreter prompt and the Predictor prompt.

\begin{figure*}[t]
    \centering
    
    \begin{subfigure}{\textwidth}
        \begin{tcolorbox}[
            colback=gray!5,
            colframe=gray!60,
            title=\textbf{Interpreter Prompt},
            arc=2mm,
            boxrule=1pt
        ]
    \small 
     You are an NLP researcher conducting an investigation into one common basis vector shared by several document embeddings. Your goal is to determine what semantic phenomenon this vector encodes—specifically, which linguistic cue, tone, word choice, narrative framing, concept, or topic is associated with high activation along this dimension.
    
    \vspace{0.5em}
    INPUT DESCRIPTION: You will be given three inputs:
    \begin{enumerate}[leftmargin=*, nosep]
        \item Max-Activating Examples — up to 15 texts that produce high activation on this feature, with activation scores.
        \item Typical-Activating Examples — up to 15 texts that activate this feature but are not necessarily among the highest activations.
        \item Zero-Activating Examples — up to 15 texts that produce zero activation on this feature.
    \end{enumerate}
    
    \vspace{0.5em}
    OUTPUT DESCRIPTION: Using the inputs, complete the following:
    \begin{enumerate}[leftmargin=*, nosep]
        \item From the \textsc{Max-Activating Examples}, list potential topics, concepts, themes, frames, and linguistic features they share. Consider multiple granularities. Give greater weight to items more prominent in higher-activation examples.
        \item Cross-check those candidates against the \textsc{Typical-Activating Examples}. Keep features that generalize across both max and typical positives; de-emphasize items that only appear in extreme cases.
        \item Using the \textsc{Zero-Activating Examples}, rule out any surviving topics/concepts/features that also appear in the zero-activating set. Be systematic.
        \item Based on steps 1–3, provide a concise, rational analysis of the cues, tone, wording, frames, concepts, or topics—at appropriate granularity—that are likely to activate this neuron. Prefer the simplest explanation that fits all evidence.
        \item Pick the single most prominent component that aligns with this vector, and express it in 4–10 words as: \texttt{FINAL:<description>}.
    \end{enumerate}
    
    \vspace{0.5em}
    Here are the max-activating examples: \texttt{<max activating examples>} \\
    Here are the typical-activating examples: \texttt{<typical activating examples>} \\
    Here are the zero-activating examples: \texttt{<zero activating examples>}
    
    \vspace{0.5em}
    Work through the steps thoroughly and analytically to interpret the vector. Output should only be in this format: \texttt{FINAL:<description>}. Do NOT return anything after these 4–10 words.
        \end{tcolorbox}
        \label{fig:prompt_step1}
    \end{subfigure}
    
    \vspace{1em} 
    
    \begin{subfigure}{\textwidth}
        \begin{tcolorbox}[
            colback=gray!5,
            colframe=gray!60,
            title=\textbf{Predictor Prompt},
            arc=2mm,
            boxrule=1pt
        ]
    \small
    You are an NLP researcher investigating one of the basis vectors of a document-embedding space. Your goal is to predict whether a given text exhibits the feature encoded by this vector.
    
    \vspace{0.5em}
    OUTPUT DESCRIPTION:
    Based on the feature description, predict whether the vector will activate for this text. If you predict activation, output 1. If not, output 0.
    
    \vspace{0.5em}
    Here is the description of the feature encoded by the vector: \texttt{<description>} \\
    Here is the text to predict: \texttt{<text>}
    
    \vspace{0.5em}
    Provide the output only in one of the following forms: \texttt{PREDICTION: 1} or \texttt{PREDICTION: 0}. Do not include anything else.
        \end{tcolorbox}
        \label{fig:prompt_step2}
    \end{subfigure}

    \caption{The prompts used for the Interpreter--Predictor Module. The Interpreter generates a semantic description of the vector, which is then fed into the Predictor to validate predictive accuracy.}
    \label{fig:full_prompts}
\end{figure*}

\section{LLM Models and Inference Settings}
\label{sec:appendix_llm_detail}
For Interpreter and Predictor LLM runs, we use open-weight, instruction-tuned Gemma~3 models: Gemma-3-27B-IT (FP8) as the Interpreter and Gemma-3-12B-IT (bfloat16) as the Predictor.\footnote{\url{https://huggingface.co/google/gemma-3-27b-it} and \url{https://huggingface.co/google/gemma-3-12b-it}.}
Both models are accessed via the DeepInfra API.\footnote{\url{https://deepinfra.com}.}
We use the smaller model for the Predictor because the task is a comparatively simple binary decision. For both Interpreter and Predictor calls, we set the maximum context length to 131{,}072 tokens.

\section{Counts of Interpretable Features Across Interpretability Score Thresholds}
\label{app:rq1_full_tables}

\pgfplotstableread[col sep=comma]{rq2_20NewsGroup_final_summary.csv}\rqTwoNews

See Table~\ref{tab:rq1-20news}, Table~\ref{tab:rq1-wos} and Table~\ref{tab:rq1-reuters}.

\begin{table}[t]
\centering
\small
\setlength{\tabcolsep}{4pt} 
\pgfplotstabletypeset[
  col sep=comma,
  string type,
  column type=c,
  columns={N,K, IS≥0.95,IS≥0.90,IS≥0.85,IS≥0.80},
  columns/N/.style={fixed, precision=1, column type=r, column name={$N$}},
  columns/K/.style={int detect, column type=r, column name={$K$}},
  columns/{IS≥0.95}/.style={int detect, column type=r, column name={IS$\ge 0.95$}},
  columns/{IS≥0.90}/.style={int detect, column type=r, column name={IS$\ge 0.90$}},
  columns/{IS≥0.85}/.style={int detect, column type=r, column name={IS$\ge 0.85$}},
  columns/{IS≥0.80}/.style={int detect, column type=r, column name={IS$\ge 0.80$}},
  every head row/.style={
    before row=\toprule,
    after row=\midrule
  },
  every last row/.style={after row=\bottomrule},
]\rqTwoNews
\caption{Number of features above interpretability score thresholds in the 20 Newsgroups dataset. IS denotes the interpretability score.}
\label{tab:rq1-20news}
\end{table}

\pgfplotstableread[col sep=comma]{rq2_webofscience_final_summary.csv}\rqTwoWOS
\begin{table}[t]
\centering
\small
\setlength{\tabcolsep}{4pt} 
\pgfplotstabletypeset[
  col sep=comma,
  string type,
  column type=c,
  columns={N,K, IS≥0.95,IS≥0.90,IS≥0.85,IS≥0.80},
  columns/N/.style={fixed, precision=1, column type=r, column name={$N$}},
  columns/K/.style={int detect, column type=r, column name={$K$}},
  columns/{IS≥0.95}/.style={int detect, column type=r, column name={IS$\ge 0.95$}},
  columns/{IS≥0.90}/.style={int detect, column type=r, column name={IS$\ge 0.90$}},
  columns/{IS≥0.85}/.style={int detect, column type=r, column name={IS$\ge 0.85$}},
  columns/{IS≥0.80}/.style={int detect, column type=r, column name={IS$\ge 0.80$}},
  every head row/.style={
    before row=\toprule,
    after row=\midrule
  },
  every last row/.style={after row=\bottomrule},
]\rqTwoWOS
\caption{Number of features above interpretability score thresholds in the Web of Science dataset. IS denotes the interpretability score.}
\label{tab:rq1-wos}
\end{table}

\pgfplotstableread[col sep=comma]{rq2_reuters_final_summary.csv}\rqTwoReuters
\begin{table}[t]
\centering
\small
\setlength{\tabcolsep}{4pt} 
\pgfplotstabletypeset[
  col sep=comma,
  string type,
  column type=c,
  columns={N,K, IS≥0.95,IS≥0.90,IS≥0.85,IS≥0.80},
  columns/N/.style={fixed, precision=1, column type=r, column name={$N$}},
  columns/K/.style={int detect, column type=r, column name={$K$}},
  columns/{IS≥0.95}/.style={int detect, column type=r, column name={IS$\ge 0.95$}},
  columns/{IS≥0.90}/.style={int detect, column type=r, column name={IS$\ge 0.90$}},
  columns/{IS≥0.85}/.style={int detect, column type=r, column name={IS$\ge 0.85$}},
  columns/{IS≥0.80}/.style={int detect, column type=r, column name={IS$\ge 0.80$}},
  every head row/.style={
    before row=\toprule,
    after row=\midrule
  },
  every last row/.style={after row=\bottomrule},
]\rqTwoReuters
\caption{Number of features above interpretability score thresholds in the Reuters dataset. IS denotes the interpretability score.}
\label{tab:rq1-reuters}
\end{table}

\section{Feature Granularity Across SAE Configurations}
\label{sec:granularity_probe_details}

\subsection{Motivation}

In the main text, RQ1 focuses on whether corpus-trained SAE features can be reliably interpreted as semantic units. As a secondary diagnostic, we also ask whether SAE configurations differ in the granularity of the validated semantic features they recover. If larger or more active SAEs refine features learned by smaller SAEs, then document-level activations from the larger SAE should more easily reconstruct the validated-feature activations of the smaller SAE than vice versa.

This analysis is not required for MonoTM, but it helps characterize how SAE configuration affects the structure of the validated feature space.

\subsection{Probe setup}

Consider two SAEs trained on the same corpus but with different configurations $(N,K)$, denoted A and B. Let $\mathcal{F}^{(A)}$ and $\mathcal{F}^{(B)}$ be the subsets of latent features that pass our label validation threshold. We use $\mathrm{IS}\ge 0.80$ throughout. Let $p = |\mathcal{F}^{(A)}|$ and $q = |\mathcal{F}^{(B)}|$, and let $M$ be the number of documents.

\subsection{Document--feature activation matrices}

For each document $d$, each SAE produces a sparse top-$K$ activation list
$\{(i, a_{d i})\}_{i\in \mathrm{TopK}(d)}$ with $a_{d i}\ge 0$. We construct a dense document--feature activation vector by accumulating these latent activations into the coordinates corresponding to validated features. For SAE A, we define $x^{(A)}_d \in \mathbb{R}_{\ge 0}^{p}$ by
\begin{equation}
x^{(A)}_{d j}
=
\sum_{i \in \mathrm{TopK}^{(A)}(d)}
a^{(A)}_{d i}\,
\mathbf{1}\!\left[i = f^{(A)}_j\right],
\end{equation}
where $(f^{(A)}_1,\dots,f^{(A)}_p)$ is a fixed ordering of $\mathcal{F}^{(A)}$. Equivalently, we keep only validated feature coordinates and set all others to zero. Stacking over documents yields $X^{(A)} \in \mathbb{R}_{\ge 0}^{M\times p}$. We analogously construct $X^{(B)} \in \mathbb{R}_{\ge 0}^{M\times q}$.

\subsection{Linear mapping probe}

We fit an affine linear map in each direction:
\begin{align}
\widehat{X}^{(B)} &= X^{(A)} W_{A\to B}^\top + \mathbf{1}\, b_{A\to B}^\top, \\
\widehat{X}^{(A)} &= X^{(B)} W_{B\to A}^\top + \mathbf{1}\, b_{B\to A}^\top,
\end{align}
where $W_{A\to B}\in\mathbb{R}^{q\times p}$ and $b_{A\to B}\in\mathbb{R}^{q}$, and symmetrically for $B\to A$. Parameters are estimated on a training split using ridge regression.\footnote{Since our goal is to quantify reconstructability rather than minimize training error, we use ridge regression to control coefficient magnitudes and reduce sensitivity to resampling, yielding a lower-variance reconstructability estimate.}

\subsection{Error metric and directional asymmetry}

Because document--feature activations are nonnegative and sparse, we evaluate reconstruction with a weighted MSE that assigns higher weights to nonzero target entries:
\begin{multline}
\mathrm{WMSE}(\widehat{Y},Y)
= \frac{\sum_{d,j} w_{dj}(\widehat{y}_{dj}-y_{dj})^2}{\sum_{d,j} w_{dj}}, \\
w_{dj} = 1 + (\lambda-1)\,\mathbf{1}[y_{dj}>0],
\end{multline}
with $\lambda=10$ in our experiments.

To make scores comparable across different target spaces, we normalize by the WMSE of a zero-predictor baseline computed on the same validation set:
\begin{equation}
\rho(A\to B)
=
\frac{\mathrm{WMSE}(\widehat{X}^{(B)},X^{(B)})}
{\mathrm{WMSE}(0,X^{(B)})}.
\end{equation}
Lower $\rho$ indicates more accurate reconstruction relative to predicting all-zero activations.

Our main statistic is the directional asymmetry
\begin{equation}
\Delta
=
\rho(\text{big}\to\text{small})
-
\rho(\text{small}\to\text{big}),
\end{equation}
where ``big'' and ``small'' refer to the ordered SAE configurations used in this probe. Negative values indicate that the larger SAE more easily reconstructs the validated-feature activation patterns of the smaller SAE than the reverse direction.

\subsection{Resampling protocol}

For each dataset, we repeat the full procedure over 10 random train/validation splits of documents. For each split, we fit both directions, $A\to B$ and $B\to A$, and compute $\rho$ on the held-out validation documents. We evaluate all ordered pairs among the four SAE configurations shown in Figure~\ref{fig:granularity}.

\subsection{Results}

\begin{table}[t]
\centering
\small
\begin{tabular}{lccc}
\toprule
Dataset & Median $\Delta$ & 95\% boot.\ CI & $\Pr(\Delta<0)$ \\
\midrule
20 Ng & $-0.111$ & $[-0.223,\;\;0.052]$ & $0.667$ \\
WoS & $-0.099$ & $[-0.164,\;-0.048]$ & $1.000$ \\
Reuters & $-0.102$ & $[-0.139,\;-0.057]$ & $1.000$ \\
\bottomrule
\end{tabular}
\caption{Summary of cross-SAE predictability asymmetry across datasets. Reported are the median $\Delta$ and a 95\% hierarchical bootstrap confidence interval over configuration pairs and random splits, along with $\Pr(\Delta<0)$.}
\label{tab:granularity_delta_summary}
\end{table}

\begin{figure*}[t]
  \centering
  \includegraphics[width=2\columnwidth]{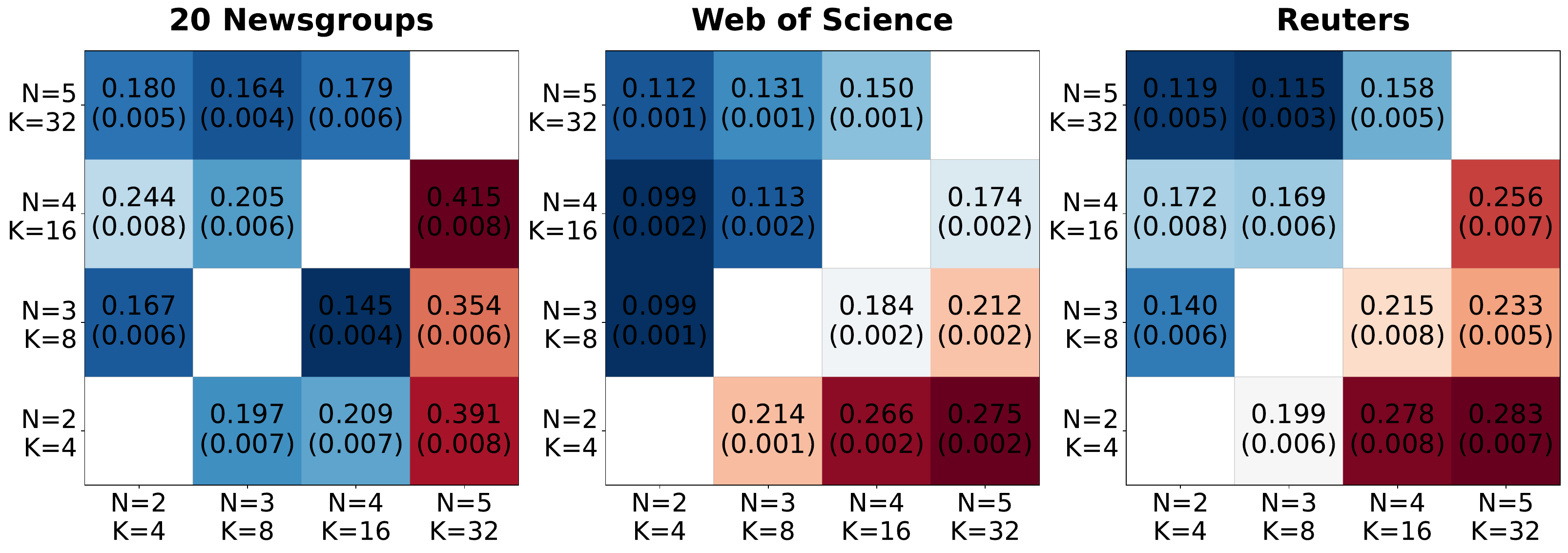}
  \caption{
  Heatmaps of linear cross-SAE reconstruction. Each cell shows the mean validation MSE ratio $\rho$ for predicting the target SAE activations from the source SAE activations, with direction shown as row $\rightarrow$ column. Parentheses give the standard deviation over 10 train/validation splits. Lower is better.
  }
\label{fig:granularity}
\end{figure*}

Figure~\ref{fig:granularity} shows directional asymmetries in cross-SAE predictability. On Web of Science and Reuters, mappings from higher-capacity configurations to lower-capacity ones consistently achieve lower normalized error than the reverse direction across essentially all configuration pairs. Aggregating across all pairs and seeds, the median $\Delta$ is negative on both datasets (Table~\ref{tab:granularity_delta_summary}), indicating that larger SAEs more easily reconstruct smaller SAEs than vice versa under this linear probe.

For 20 Newsgroups, the aggregate median $\Delta$ is also negative but less conclusive under resampling. The asymmetry is strongest when comparing the most separated configurations, $(N{=}2,K{=}4)$ and $(N{=}5,K{=}32)$, consistent with a larger gap in representational resolution.

\subsection{Interpretation and limitations}

Taken together, these results are consistent with a granularity shift as SAE capacity and activity increase. Features learned in larger SAEs appear to contain sufficient information to linearly reconstruct the validated-feature activations of smaller SAEs, while the reverse reconstruction is systematically harder.

This probe does not establish one-to-one feature correspondences, and it may underestimate nonlinear relationships between feature spaces. It should therefore be interpreted as a diagnostic of representational resolution rather than as direct evidence that individual features split cleanly across SAE configurations.

\section{Document--Topic Mixture Evaluation Details}
\label{sec:eval_details}

We evaluate document--topic mixture quality through alignment with gold label topics. This evaluation is intended to measure whether the inferred topic mixtures recover the benchmark category structure, rather than whether topic descriptors form coherent word lists.

For 20 Newsgroups and Web of Science, which are single-label datasets, we assign each document to its highest-probability inferred topic,
\[
\hat{z}_d = \arg\max_k \theta_{d,k}.
\]
We then compute the confusion matrix between inferred topics and gold labels and use the Hungarian algorithm to find the one-to-one topic--label matching that maximizes total agreement. After applying this mapping, we treat the aligned topic assignments as multi-class predictions and report Micro-F1 and Macro-F1.

Reuters is multi-label and highly imbalanced, so we evaluate it as a multi-label topic--label alignment problem. For each candidate threshold
\[
t \in \{0.1, 0.2, 0.3, 0.4, 0.5\},
\]
we predict topic $k$ for document $d$ when $\theta_{d,k} > t$. We then compute pairwise F1 scores between each gold label topic and each inferred topic, use the Hungarian algorithm to obtain a one-to-one label--topic matching, and compute Micro-F1 and Macro-F1 over the matched pairs. We apply the same threshold grid to all models and report each metric at its best threshold over this grid. Documents with no Reuters gold label are retained in the evaluation, so a model can avoid false positives on these documents only by assigning all topic probabilities below the selected threshold.

\section{Full SAE Hyperparameter Heatmaps for Mixture Estimation}
\label{sec:rq2_full_heatmaps}

\begin{figure*}[t]
  \centering
  \includegraphics[width=\textwidth]{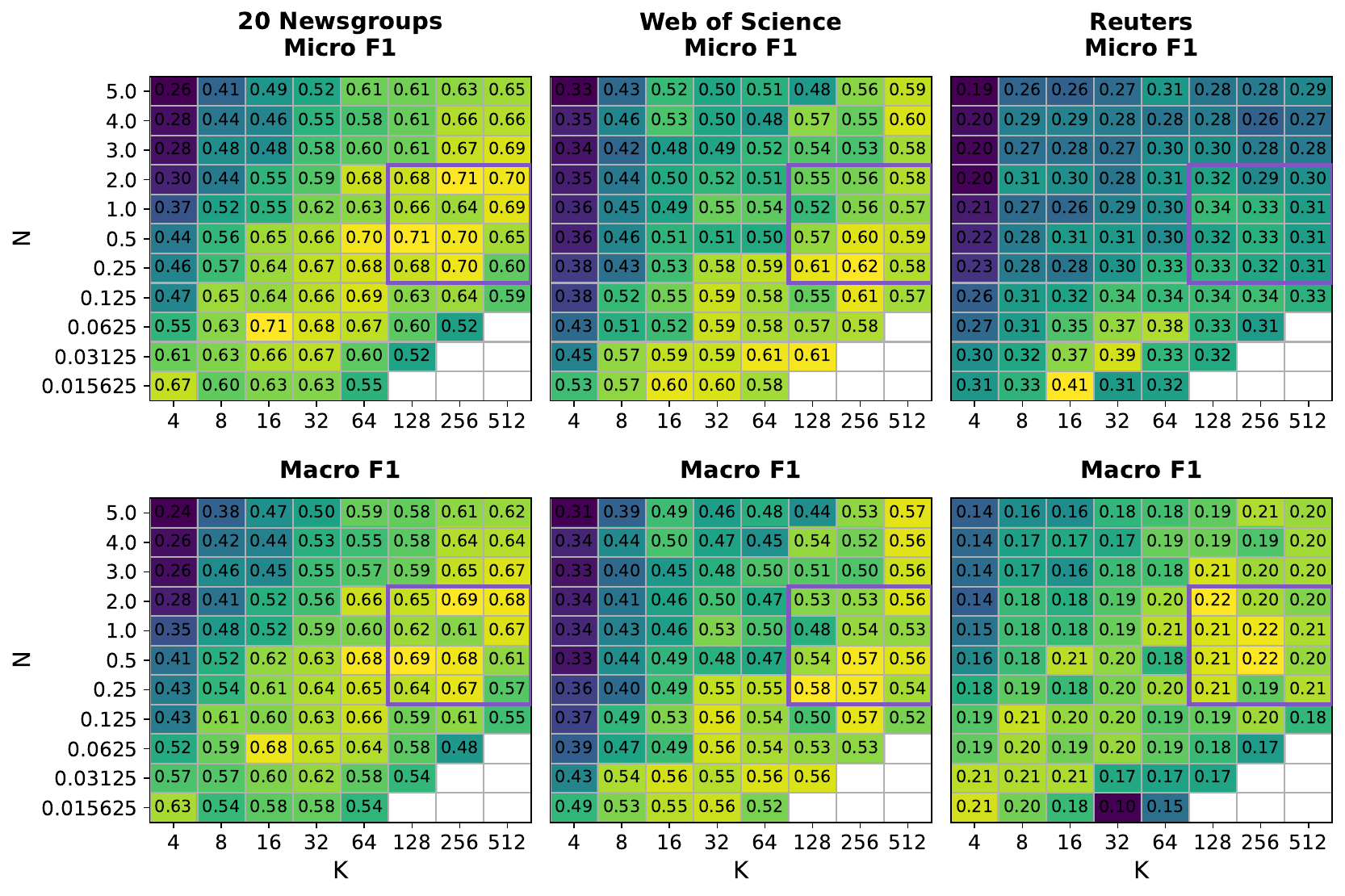}
  \caption{Heatmaps of Micro-F1 and Macro-F1 scores over the full SAE hyperparameter grid for BoF+LDA. The purple box indicates the $(N,K)$ region whose values are averaged to produce the scores reported in Table~\ref{tab:classification_results}.}
  \label{fig:rq2heatmaps}
\end{figure*}

\begin{figure}[t]
  \centering
  \includegraphics[width=1\columnwidth]{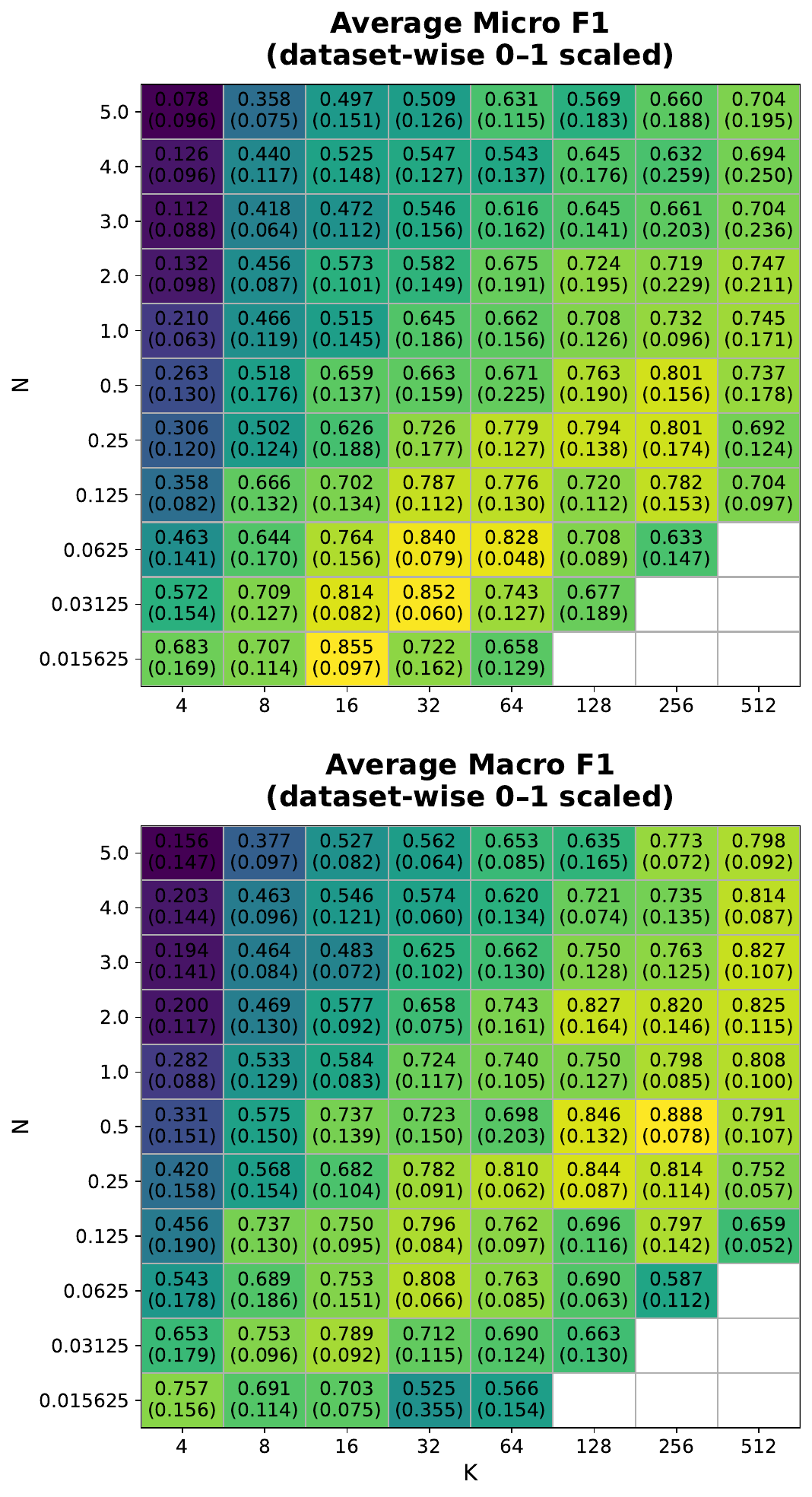}
  \caption{Heatmaps of Micro-F1 (top) and Macro-F1 (bottom) scores, shown as mean (standard deviation) over three datasets, computed from 9 data points for each $(N, K)$. F1 scores were scaled to the $[0,1]$ range on a per-dataset basis to prevent any single dataset from dominating performance.}
  \label{fig:scaledhm}
\end{figure}

Figure~\ref{fig:rq2heatmaps} reports Micro-F1 and Macro-F1 over the full SAE hyperparameter grid for BoF+LDA on each dataset. Figure~\ref{fig:scaledhm} provides a complementary aggregate view by averaging performance across datasets after scaling each dataset's F1 scores to the $[0,1]$ range. Together, these figures support the hyperparameter-selection discussion in Section~\ref{sec:obsB_theta_scale} by showing how document--topic mixture quality varies with dictionary capacity $N$ and per-document activity $K$.

Across datasets, two patterns are visible. First, very wide SAEs paired with very small $K$ tend to perform poorly. In this regime, the dictionary has high capacity, but each document is allowed to activate only a small number of features. The resulting BoF representation is therefore too constrained: it exposes many possible pseudo-token types, but provides too little per-document evidence for stable topic inference. Second, boundary cases where $mN=K$ are also weak. Since the top-$K$ operator retains all available units in this case, the model no longer imposes a meaningful sparsity bottleneck, which appears to reduce the usefulness of SAE latents as topic-modeling signals.

By contrast, stronger performance appears in a balanced scaling band where $N$ and $K$ increase together. These configurations preserve an effective sparsity constraint while allowing each document to express a richer set of SAE features. This pattern is visible in the per-dataset heatmaps in Figure~\ref{fig:rq2heatmaps} and remains apparent in the dataset-aggregated heatmaps in Figure~\ref{fig:scaledhm}. The purple region in Figure~\ref{fig:rq2heatmaps} marks the robust subset of this balanced regime used to summarize BoF+LDA in Table~\ref{tab:classification_results}. We use this region rather than a single best cell to avoid overemphasizing dataset- or seed-specific hyperparameter choices.

\section{Document--Topic Estimation Using Only Interpretable Features}
\label{sec:appendix}
\begin{figure*}[t]
  \centering
  \includegraphics[width=\textwidth]{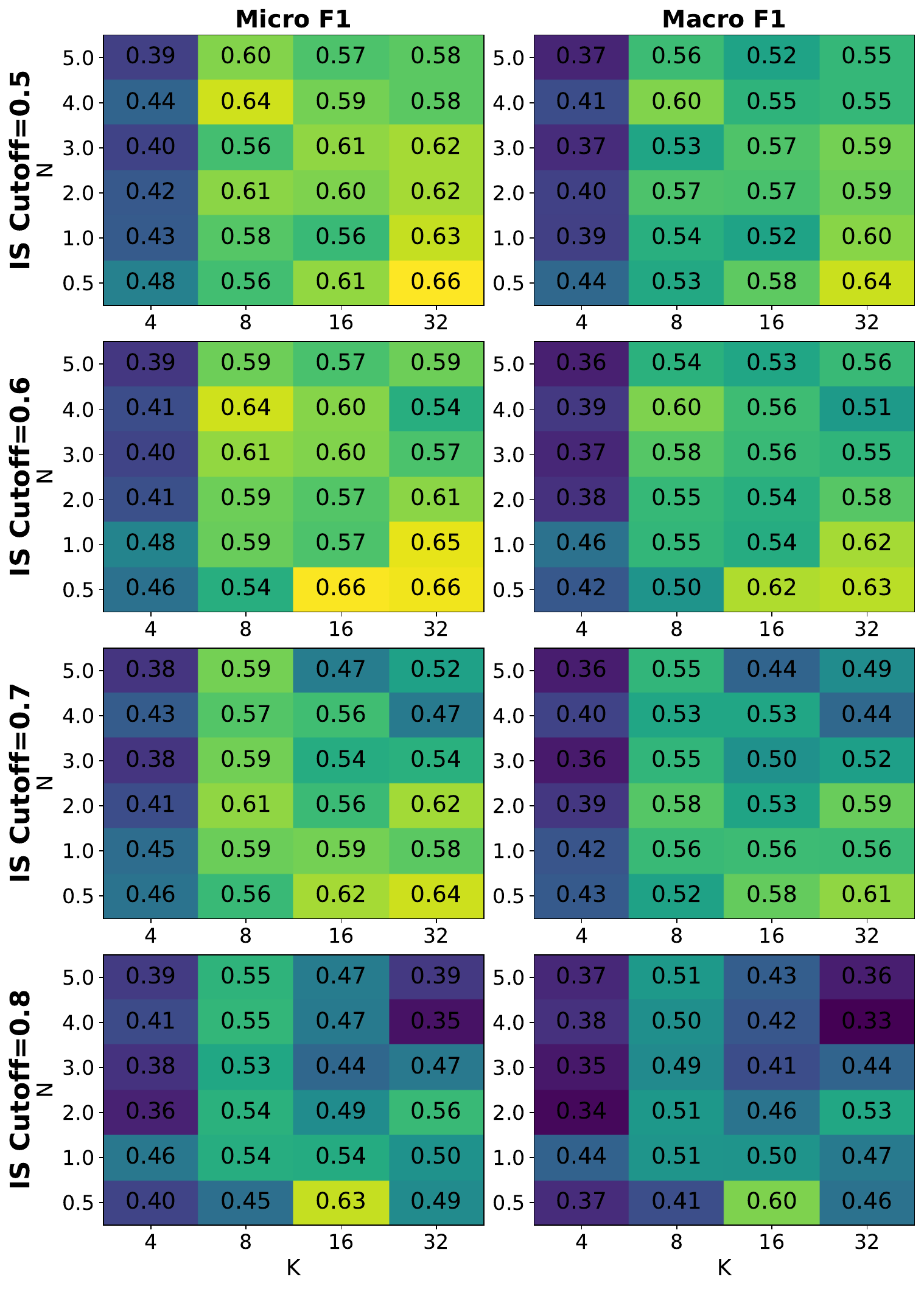}
    \caption{Heatmaps of micro-F1 and macro-F1 scores for document–topic mixtures on the 20 Newsgroups dataset, inferred using only features with interpretability scores (IS) above various cutoffs.
    }
  \label{fig:20ngf1fiter}
\end{figure*}

\begin{figure*}[t]
  \centering
  \includegraphics[width=\textwidth]{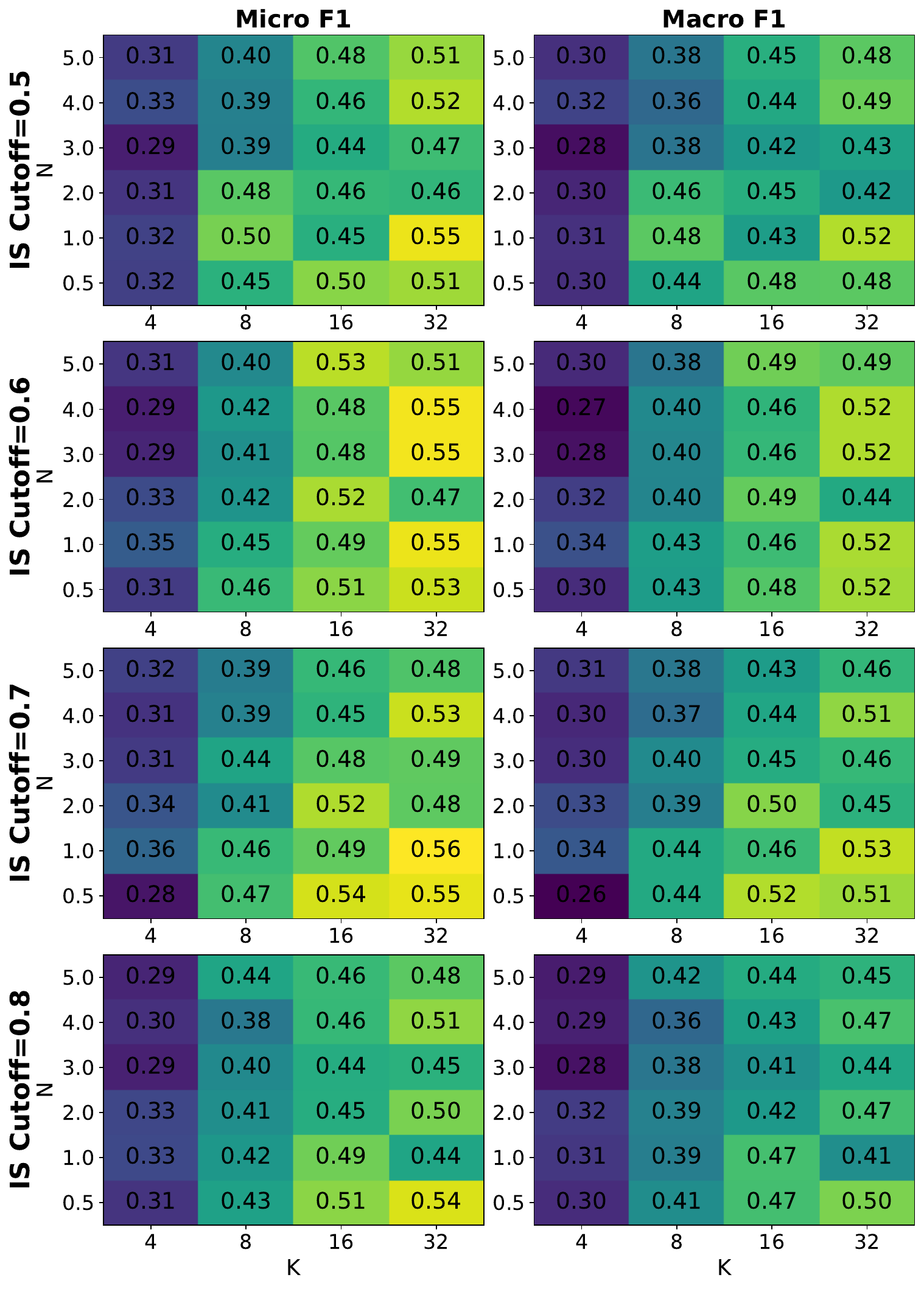}
    \caption{Heatmaps of micro-F1 and macro-F1 scores for document–topic mixtures on the Web of Science dataset, inferred using only features with interpretability scores (IS) above various cutoffs.
    }
  \label{fig:wosf1fiter}
\end{figure*}

\begin{figure*}[t]
  \centering
  \includegraphics[width=\textwidth]{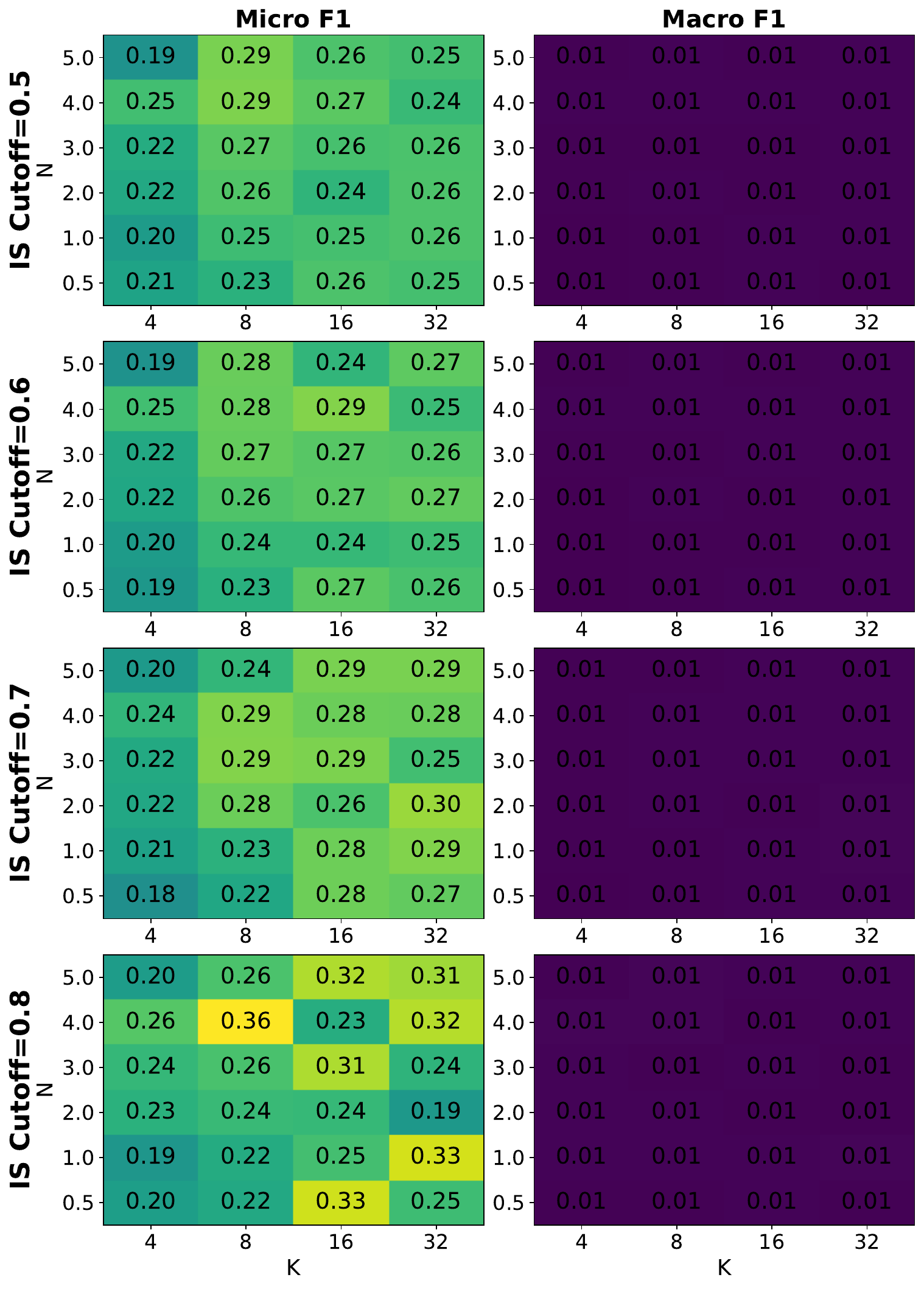}
    \caption{Heatmaps of micro-F1 and macro-F1 scores for document–topic mixtures on the Reuters dataset, inferred using only features with interpretability scores (IS) above various cutoffs. To evaluate against the multi-label gold labels, we treat topics whose inferred document–topic probabilities exceed 0.2 as predicted classes.
    }
  \label{fig:reutersf1fiter}
\end{figure*}

See Figure~\ref{fig:20ngf1fiter}, Figure~\ref{fig:wosf1fiter}, and Figure~\ref{fig:reutersf1fiter}.

\section{Sample Topic--Feature Representations}
\label{sec:sample-MonoTM-output}

Table \ref{tab:all_datasets_features} presents a subset of MonoTM's topic representation output, reporting the top-5 labeled features per topic for three topics in each dataset.

We use an interpretable-feature SAE with $(N{=}1, K{=}8)$ and filter features by $\mathrm{IS}(f)\ge 0.8$. This is not the richest configuration we tested: larger settings, such as $(N{=}5, K{=}32)$, yield substantially more interpretable features. However, the top feature labels for each topic show that even with interpretable features obtained from a relatively small SAE, the feature labels associated with high $\beta$ values align well with the gold topic labels and provide a finer-grained understanding of the corpus than topic labels alone.

\begin{table*}[t]
\centering
\footnotesize
\begin{tabular}{llp{8.8cm}r}
\toprule
Dataset & Topic & Feature label & $\beta$ \\
\midrule
20 Newsgroups & rec.sport.hockey & NHL hockey game and player information & .402 \\
20 Newsgroups & rec.sport.hockey & Sports scores reported on Usenet newsgroups & .048 \\
20 Newsgroups & rec.sport.hockey & Frustrated fan complaints about sports broadcasting decisions & .038 \\
20 Newsgroups & rec.sport.hockey & Passionate sports discussion in online forums & .031 \\
20 Newsgroups & rec.sport.hockey & Sports stats requests and early internet culture & .030 \\
\midrule
20 Newsgroups & soc.religion.christian & Intense Christian theological debate and interpretation & .348 \\
20 Newsgroups & soc.religion.christian & Religious debate, interpretation, and personal belief defense & .061 \\
20 Newsgroups & soc.religion.christian & Traditional Catholic theology and liturgical concerns & .041 \\
20 Newsgroups & soc.religion.christian & Skepticism towards religious belief and dogma & .040 \\
20 Newsgroups & soc.religion.christian & Christian theological discussion and apologetic reasoning & .030 \\
\midrule
20 Newsgroups & rec.motorcycles & Motorcycle riding advice and safety discussion & .481 \\
20 Newsgroups & rec.motorcycles & Motorcycle countersteering technique and related debate & .035 \\
20 Newsgroups & rec.motorcycles & Discussions of drunk driving and motorcycles & .033 \\
20 Newsgroups & rec.motorcycles & Motorcycle culture, camaraderie, road-based social signaling & .032 \\
20 Newsgroups & rec.motorcycles & Motorcycle enthusiast discussion: ownership, specs, and banter & .031 \\
\midrule
\addlinespace[2pt]
Web of Science & Medical & Cardiovascular disease research and clinical information & .016 \\
Web of Science & Medical & Gastrointestinal function, microbiome, and symptomology & .016 \\
Web of Science & Medical & Bladder dysfunction, urological assessment, and clinical research & .015 \\
Web of Science & Medical & Osteoporosis research, bone density, and skeletal health & .014 \\
Web of Science & Medical & Neurobiological basis of psychiatric disease research & .013 \\
\midrule
Web of Science & Psychology & Internal psychological states and research analysis & .029 \\
Web of Science & Psychology & Parenting challenges, family stress, child well-being & .025 \\
Web of Science & Psychology & Social judgment and character attribution processes & .020 \\
Web of Science & Psychology & Psychiatric comorbidity, distress, and vulnerability factors & .019 \\
Web of Science & Psychology & Autism spectrum disorder and related neurodevelopmental features & .017 \\
\midrule
Web of Science & CS & Advanced computer and network security threat analysis & .030 \\
Web of Science & CS & Computer vision algorithm development and evaluation & .023 \\
Web of Science & CS & Parallel computing, performance optimization, technical implementation details & .022 \\
Web of Science & CS & Cryptographic security and data privacy techniques & .019 \\
Web of Science & CS & Machine learning methodology and technical rigor & .016 \\
\midrule
\addlinespace[2pt]
Reuters & earn & Comparative financial performance reporting, earnings vs prior period & .516 \\
Reuters & earn & Comparative financial performance, ``X vs Y'' reporting & .175 \\
Reuters & earn & Quarterly financial report key metrics comparison & .073 \\
Reuters & earn & Financial turnaround, profit versus loss reporting & .063 \\
Reuters & earn & Corporate quarterly financial results reporting and metrics & .054 \\
\midrule
Reuters & acq & Corporate acquisition and ownership transfer announcements & .311 \\
Reuters & acq & Corporate mergers and acquisitions announcements, financial details & .089 \\
Reuters & acq & Mergers, acquisitions, and corporate restructuring events & .088 \\
Reuters & acq & Corporate merger and acquisition announcements (LOI) & .081 \\
Reuters & acq & Energy sector financial deal reporting details & .051 \\
\midrule
Reuters & money-fx & Central bank money market interventions and rates & .366 \\
Reuters & money-fx & Bank of England money market liquidity revisions & .227 \\
Reuters & money-fx & Central bank money market intervention reporting & .097 \\
Reuters & money-fx & Bank of England liquidity intervention reporting & .075 \\
Reuters & money-fx & British monetary policy and currency market signals & .033 \\
\bottomrule
\end{tabular}
\caption{Part of the final MonoTM output showing topic--feature distributions. The table lists the top five validated feature labels for three topics from each dataset. Gold labels from the original datasets are assigned using the alignment procedure in Section~\ref{sec:rq2_results} and are shown only for orientation. $\beta$ denotes the probability of each feature within the topic--feature distribution.}
\label{tab:all_datasets_features}
\end{table*}

\section{Full MonoTM Algorithm and Inference Details}
\label{sec:MonoTM-details}

This appendix expands the MonoTM algorithm summarized in Section~\ref{sec:MonoTM-algorithm}. Let $\mathcal{D}=\{1,\dots,D\}$ index documents and let $T$ be the number of topics. MonoTM estimates document--topic mixtures from a full SAE bag-of-features representation, then estimates a topic--feature distribution over validated interpretable SAE features with the mixtures held fixed.

We allow the SAE used for mixture estimation and the SAE used for interpretation to be different. We denote their activations by $\tilde h^{\mathrm{mix}}$ and $\tilde h^{\mathrm{int}}$, respectively. The single-SAE case is recovered by setting these two representations to be the same. For each document $d$, let $\theta_d\in\Delta^T$ denote its document--topic mixture, and let $\Theta\in\mathbb{R}_{\ge0}^{D\times T}$ collect these mixtures row-wise. Let $\mathcal{F}_\tau=\{1,\dots,V\}$ denote the set of validated interpretable SAE features used for topic description. Each feature $f\in\mathcal{F}_\tau$ has a natural-language label $\ell_f$ and interpretability score $\mathrm{IS}(f)$. Our goal is to estimate a topic--feature matrix $B\in\mathbb{R}_{\ge0}^{T\times V}$ whose row $\beta_t\in\Delta^V$ ranks validated interpretable features for topic $t$.

\paragraph{Stage 1: estimating document--topic mixtures.}
For mixture estimation, MonoTM uses the full mixture-SAE representation rather than only validated features. For a top-$K$ SAE with expansion factor $N$, each document embedding yields a sparse nonnegative activation vector $\tilde h^{\mathrm{mix}}_d\in\mathbb{R}_{\ge0}^{mN}$ with at most $K$ nonzero entries. We construct a document--feature matrix $X^{\mathrm{mix}}\in\mathbb{R}_{\ge0}^{D\times mN}$ by setting
\begin{equation}
X^{\mathrm{mix}}_{d,j}=\tilde h^{\mathrm{mix}}_{d,j},
\label{eq:MonoTM-full-bof}
\end{equation}
and fit LDA with $T$ topics on $X^{\mathrm{mix}}$ to obtain $\Theta$. This stage uses all active SAE features because Section~\ref{sec:filter} shows that strict interpretability filtering generally hurts mixture estimation.

\paragraph{Stage 2: constructing the interpretable feature matrix.}
For topic description, MonoTM restricts to interpretation-SAE features that (i) activate in at least 30 documents, (ii) have a nonempty natural-language label, and (iii) satisfy $\mathrm{IS}(f)\ge\tau$, with $\tau=0.8$ in our experiments. The interpretable document--feature matrix $C\in\mathbb{R}_{\ge0}^{D\times V}$ is defined by
\begin{equation}
c_{d,f}=\max(\tilde h^{\mathrm{int}}_{d,f},0)\cdot \mathrm{IS}(f),
\qquad f\in\mathcal{F}_\tau.
\label{eq:f1_weighted_features}
\end{equation}
This weighting makes the final topic descriptors emphasize features with stronger validation evidence.

\paragraph{Stage 3: estimating topic--feature distributions.}
Given $\Theta$ and $C$, MonoTM estimates $B$ by maximizing the fixed-$\Theta$ mixture log-likelihood
\begin{equation}
\mathcal{L}(B)=
\sum_{d=1}^{D}\sum_{f=1}^{V}
c_{d,f}\log\Big(\sum_{t=1}^{T}\theta_{d,t}\beta_{t,f}\Big),
\label{eq:MonoTM-fixed-theta-ll}
\end{equation}
subject to $\beta_t\in\Delta^V$ for all $t$. After estimating $B$, topic $t$ is represented by its ranked feature distribution $\beta_t$; the reported descriptor list consists of the top-ranked labels $\ell_f$ under $\beta_{t,f}$, with associated probabilities.

\subsection{Dirichlet-smoothed EM}
\label{sec:em}

We optimize Eq.~\eqref{eq:MonoTM-fixed-theta-ll} with an EM algorithm that treats latent topic assignments as missing data. Let
\begin{equation}
q_{d,f,t}=p(z=t\mid d,f)
=\frac{\theta_{d,t}\beta_{t,f}}{\sum_{t'}\theta_{d,t'}\beta_{t',f}}.
\label{eq:posterior}
\end{equation}
The E-step computes expected topic--feature counts:
\begin{equation}
N_{t,f}=\sum_{d=1}^{D} c_{d,f}\,q_{d,f,t}.
\label{eq:Nkf}
\end{equation}
The M-step updates $B$ with symmetric Dirichlet smoothing $\eta>0$:
\begin{equation}
\beta_{t,f}=\frac{N_{t,f}+\eta}{\sum_{f'}(N_{t,f'}+\eta)}.
\label{eq:mstep}
\end{equation}
We iterate until relative improvement in Eq.~\eqref{eq:MonoTM-fixed-theta-ll} falls below a tolerance. To avoid zero-probability features and stabilize topics when $C$ is sparse, we use the data-dependent heuristic
\begin{equation}
\eta=\frac{1}{T}+\frac{\bar L}{V},
\end{equation}
where $\bar L$ is the average number of nonzero validated features per document.

\subsection{Using Separate Mixture and Interpretation SAEs}
\label{sec:decoupling}

MonoTM does not require a single SAE to simultaneously optimize mixture estimation and interpretability. Even when a single SAE is used end-to-end, we do not interpret topics using the topic--feature distribution implicitly produced during $\Theta$ inference. Instead, topic interpretation is based on a separately estimated topic--feature distribution over validated interpretable features. When the SAE configuration that yields the most useful validated feature set is suboptimal for estimating $\Theta$, a user can train a second SAE optimized for mixture quality while using an interpretability-optimized SAE to construct $C$.

\section{Excluded-feature audit details}
\label{sec:excluded-feature-audit-details}

This appendix gives the full procedure for the excluded-feature audit in Section~\ref{sec:excluded-feature-audit}. The audit asks whether features excluded from the final MonoTM descriptor vocabulary nevertheless contain topic-defining semantic information that is absent from the validated descriptor set.

\paragraph{Configurations.}
For all datasets, the document--topic mixture matrix $\Theta$ is the same one used in the main RQ3 analysis. It is estimated from the mixture-quality SAE with expansion factor $N=0.5$ and top-$K=256$, using seed 123. The number of topics is fixed to the number of gold categories: $T=20$ for 20 Newsgroups, $T=7$ for Web of Science, and $T=47$ for Reuters. The interpretation-side SAE used for the audit has $N=2$ and $K=16$.

\paragraph{All-active interpretation feature matrix.}
Let $\mathcal{F}_{\mathrm{act}}$ denote the set of interpretation-SAE features that activate in at least 30 documents, and let $\ell_f$ be the generated label for feature $f$. The validated descriptor set used by MonoTM is
\[
\mathcal{F}_{\mathrm{val}}
=\{f\in\mathcal{F}_{\mathrm{act}}: \mathrm{IS}(f)\ge .8 \text{ and } \ell_f\neq \emptyset\}.
\]
For the audit, we instead construct an all-active feature matrix $A\in\mathbb{R}_{\ge0}^{D\times |\mathcal{F}_{\mathrm{act}}|}$ with entries
\[
a_{d,f}=\max(\tilde h_{d,f},0),
\]
where $\tilde h_{d,f}$ is the gain-adjusted top-$K$ activation of the interpretation SAE. Unlike the final descriptor estimation step, the audit does not weight feature activations by $\mathrm{IS}(f)$. This avoids building the conclusion into the measurement: interpretability-score weighting would mechanically reduce the mass of lower-validation features. Document rows are aligned between $\Theta$, the interpretation-SAE activations, and raw documents using saved row IDs.

\paragraph{All-feature audit distribution.}
Using fixed $\Theta$ and the raw all-active matrix $A$, we estimate an auxiliary topic--feature distribution $B^{\mathrm{all}}$ by maximizing
\[
\mathcal{L}(B^{\mathrm{all}})
=\sum_{d=1}^{D}\sum_{f\in\mathcal{F}_{\mathrm{act}}}
 a_{d,f}
 \log\left(\sum_{t=1}^{T}\theta_{d,t}\beta^{\mathrm{all}}_{t,f}\right),
\]
subject to $\beta^{\mathrm{all}}_t\in\Delta^{|\mathcal{F}_{\mathrm{act}}|}$ for every topic $t$. We use the same Dirichlet-smoothed EM routine as in the fixed-$\Theta$ MonoTM descriptor step. This distribution is used only for auditing; the final MonoTM descriptors are still estimated over the validated feature vocabulary.

\paragraph{High-risk excluded set.}
For each topic $t$, all active features are ranked by $\beta^{\mathrm{all}}_{t,f}$. We define
\[
\operatorname{rank}^{\mathrm{all}}_t(f)
=1+\left|\{f':\beta^{\mathrm{all}}_{t,f'}>\beta^{\mathrm{all}}_{t,f}\}\right|.
\]
For each feature $f$ with $\mathrm{IS}(f)<.8$, let $t_f^\star$ be the topic where it has its best rank. The high-risk excluded set is
\[
\mathcal{R}_{20}
=\{(t_f^\star,f): \mathrm{IS}(f)<.8,\; \operatorname{rank}^{\mathrm{all}}_{t_f^\star}(f)\le20\}.
\]
This set contains lower-validation features that are sufficiently associated with at least one topic to be plausible missing descriptors.

\paragraph{Coverage test.}
For each $(t,f)\in\mathcal{R}_{20}$, we compare $f$ with validated same-topic descriptors. The candidate pool is
\[
\mathcal{V}_{t,M}
=\{g\in\mathcal{F}_{\mathrm{val}}:\operatorname{rank}^{\mathrm{all}}_t(g)\le M\}.
\]
We report $M=50$ in the main text and include $M=20$ and $M=100$ as sensitivity settings. The $M=20$ setting asks whether the feature is covered by the most compact validated descriptor region; $M=50$ asks whether it is covered by the broader high-ranked validated descriptor distribution; $M=100$ checks sensitivity to a wider candidate set.

Let $a_f$ be the document-level activation vector for feature $f$. We compute
\[
\cos_{\mathrm{act}}(f,g)
=\frac{a_f^\top a_g}{\|a_f\|_2\|a_g\|_2}.
\]
The nearest validated same-topic descriptor is
\[
g^\star(f,t)=\arg\max_{g\in\mathcal{V}_{t,M}}\cos_{\mathrm{act}}(f,g).
\]
For comparison, we sample random validated descriptors from the same candidate pool and compute the random-baseline activation cosine. Table~\ref{tab:excluded-feature-coverage-details} reports the aggregate results.

\begin{table*}[t]
\centering
\small
\begin{tabular}{llrrrrr}
\toprule
Dataset & Candidate pool & $n$ & Med. nearest & Med. random & Med. $\Delta$ & $\%\cos_{\mathrm{act}}\ge .2$ \\
\midrule
20NG & top-20 & 129 & .239 & .072 & .154 & 60.5 \\
20NG & top-50 & 129 & .295 & .055 & .218 & 72.9 \\
20NG & top-100 & 129 & .312 & .044 & .262 & 78.3 \\
WoS & top-20 & 26 & .360 & .104 & .235 & 76.9 \\
WoS & top-50 & 26 & .419 & .079 & .295 & 88.5 \\
WoS & top-100 & 26 & .442 & .066 & .349 & 96.2 \\
Reuters & top-20 & 334 & .170 & .059 & .105 & 41.6 \\
Reuters & top-50 & 334 & .170 & .043 & .130 & 45.2 \\
Reuters & top-100 & 334 & .188 & .029 & .153 & 47.9 \\
\bottomrule
\end{tabular}
\caption{Coverage audit details for lower-validation high-association features. The high-risk excluded set is fixed as features with $\mathrm{IS}(f)<.8$ and best all-feature topic rank at most 20. Candidate pool size controls how many validated same-topic descriptors are allowed to provide coverage. Median random cosine is computed from random validated descriptors sampled from the same topic-specific candidate pool.}
\label{tab:excluded-feature-coverage-details}
\end{table*}

\paragraph{Heuristic label-type flags.}
We also assign a coarse heuristic label type to each excluded feature based only on keywords in its generated feature label. These flags are not manual annotations and are not used as a validation metric. They are included to characterize broad patterns, especially the difference between Reuters and the other datasets.

Labels are flagged as \textit{format/register/numeric} if they include terms related to document form, boilerplate, reporting style, technical artifacts, numerical data, or financial-news announcements, including keywords such as: \textit{format, boilerplate, template, placeholder, metadata, header, posting, email, message, announcement, reporting, reports, quarterly, qtly, dividend, earnings, record date, subscription, unsubscribe, FAQ, manual, instructions, technical support, code, script, program, software package, release, version, archive, file, conversion, table, tabular, numerical, numeric, quantitative, percentage, percent, price, market data, Reuter, newswire}. Labels are flagged as \textit{broad discourse} if they include terms related to debate, critique, controversy, policy, institutional or political discussion, economic conflict, or broad framing, including keywords such as: \textit{discussion, debate, critique, controversy, tensions, policy, institutional, economic, financial, trade, political, rhetoric, argument, concerns, issues, context, general, broad}. Remaining labels are assigned the default \textit{semantic} flag. The rules are applied in this order: format/register/numeric first, broad discourse second, and semantic as the default.

\paragraph{Examples from the excluded-feature audit.}

Table~\ref{tab:excluded-feature-examples} gives deterministically selected examples from the label-type audit. To avoid cherry-picking, examples are selected by a fixed rule: within each dataset and heuristic label-type category, we sort features by best all-feature topic rank, break ties by larger $\beta^{\mathrm{all}}_{t,f}$, and show the first available feature. The table is intended to make the heuristic categories transparent, not to serve as a separate evaluation.

\begin{table*}[t]
\centering
\footnotesize
\begin{tabular}{llrp{4.5cm}p{4.5cm}r}
\toprule
Dataset & Type & $\mathrm{IS}(f)$ & Excluded label & Nearest validated label & $\cos_{\mathrm{act}}$ \\
\midrule
20NG & semantic & .410 & Patient narratives of challenging medical experiences & Alternative medicine debate and medical authority skepticism & .337 \\
20NG & broad & .783 & Detailed, analytical discussion of professional hockey & Toronto Maple Leafs playoff game discussion \& fandom & .301 \\
20NG & format/reg. & .727 & Sports news and scores reporting via Usenet & Ironic, irreverent, and self-aware internet discourse & .176 \\
WoS & semantic & .333 & Rainwater harvesting for water resource management & Rainwater harvesting for sustainable water management & .732 \\
WoS & broad & .750 & Distributed computing system architecture and design discuss & Intelligent interconnected devices and adaptive environments & .443 \\
WoS & format/reg. & .750 & Novel materials for photocatalytic energy conversion & Photocatalysis, solar energy, material science research & .692 \\
Reuters & semantic & .667 & Bank ownership stake acquisitions and control & Corporate deals, mergers, and joint ventures reporting & .341 \\
Reuters & broad & .769 & International trade disputes and policy tensions & International trade disputes and government intervention & .315 \\
Reuters & format/reg. & .256 & US-USSR grain trade reporting, quantitative data & U.S. farm policy and commodity support programs & .312 \\
\bottomrule
\end{tabular}
\caption{Deterministically selected examples from the excluded-feature label-type audit. Examples are selected by sorting, within each dataset and heuristic type, by best all-feature topic rank and then by larger audit-distribution topic mass.}
\label{tab:excluded-feature-examples}
\end{table*}

Concrete high-association cases illustrate the two main patterns. In 20 Newsgroups, an excluded feature labeled ``Israeli policy critique with Nazi analogies'' is ranked first under $B^{\mathrm{all}}$ for the topic aligned to \texttt{talk.politics.mideast}. Its nearest validated same-topic descriptor is ``Israeli-Palestinian conflict, criticism, and political dispute,'' with activation cosine $.650$. In Web of Science, the excluded feature ``Rainwater harvesting for water resource management'' is ranked first for the Civil topic, but its nearest validated same-topic descriptor is ``Rainwater harvesting for sustainable water management,'' with activation cosine $.732$. These cases look less like independent missing topics and more like lower-validation variants of semantic regions already covered by validated descriptors. By contrast, Reuters features such as ``Financial dividend and earnings report announcements'' or ``Dividend and earnings financial report language'' are systematic and topic-associated, but they primarily describe reporting format or financial-news register rather than substantive topical content.

\section{Descriptor Baseline and Topic-Naming Details}
\label{sec:descriptor-baseline-details}

For the controlled descriptor comparison in Section~\ref{sec:descriptor-comparison}, all methods describe the same fixed document--topic mixtures $\Theta$. We generate descriptors for each inferred topic using four mechanisms: $\theta$-weighted TF-IDF terms, BERTopic-style c-TF-IDF terms, free-form LLM-generated descriptors, and MonoTM validated SAE-feature descriptors.

\paragraph{Word-based descriptor preprocessing.}
Both word-based baselines are implemented with \texttt{scikit-learn} vectorizers. We do not apply stemming or lemmatization. The vectorizers lowercase all text, remove English stopwords, and use unigrams only. We use the token pattern
\[
\texttt{(?u)\textbackslash b[a-zA-Z][a-zA-Z0-9\_\textbackslash -]\{2,\}\textbackslash b},
\]
which keeps tokens that begin with an ASCII letter and have length at least three, while allowing later alphanumeric, underscore, and hyphen characters. This removes many short tokens, numbers-only tokens, and punctuation artifacts. We use a maximum vocabulary size of 50,000 terms.

\paragraph{$\theta$-weighted TF-IDF.}
For the $\theta$-weighted TF-IDF baseline, we fit a \texttt{TfidfVectorizer} on the full aligned corpus with \texttt{lowercase=True}, \texttt{stop\_words=english}, \texttt{ngram\_range=(1,1)}, \texttt{min\_df=2}, \texttt{max\_df=0.95}, \texttt{max\_features=50000}, \texttt{norm=None}, and \texttt{sublinear\_tf=True}. For each topic $t$, let $D_t^{50}$ be the 50 documents with the largest $\theta_{d,t}$. We normalize topic weights within this top-document set,
\[
\bar{\theta}_{d,t}
=
\frac{\theta_{d,t}}
{\sum_{d' \in D_t^{50}} \theta_{d',t}},
\]
and score each term $w$ by
\[
s_t^{\mathrm{tfidf}}(w)
=
\sum_{d \in D_t^{50}}
\bar{\theta}_{d,t}
\operatorname{tfidf}_{d,w}.
\]
We report the 30 highest-scoring terms for each topic.

\paragraph{c-TF-IDF.}
For the c-TF-IDF baseline, we form one class document for each topic by concatenating the same 50 top-$\theta$ documents. We then fit a \texttt{CountVectorizer} with \texttt{lowercase=True}, \texttt{stop\_words=english}, \texttt{ngram\_range=(1,1)}, \texttt{min\_df=1}, \texttt{max\_features=50000}, and the same token pattern as above. Let $n_{t,w}$ be the count of term $w$ in topic class document $t$, and let $L_t=\sum_w n_{t,w}$ be the class-document length. We compute
\[
\operatorname{tf}_{t,w}
=
\frac{n_{t,w}}{L_t}.
\]
Let $\operatorname{df}_{\mathrm{class}}(w)$ be the number of topic class documents in which $w$ appears, and let $\bar{L}$ be the average class-document length across topics. We use the class-based inverse-document-frequency term
\[
\operatorname{idf}_{w}
=
\log\left(1 + \frac{\bar{L}}
{\max(\operatorname{df}_{\mathrm{class}}(w),1)}\right),
\]
and rank terms by
\[
s_t^{\mathrm{ctfidf}}(w)
=
\operatorname{tf}_{t,w}
\operatorname{idf}_w.
\]
We report the 30 highest-scoring terms for each topic.

\paragraph{Free-form LLM descriptor baseline.}
The LLM summary baseline is intended to be a strong post-hoc descriptor baseline. For each topic, the prompt receives 50 topic-associated documents. The 25 highest-$\theta$ documents are always included, and the remaining 25 documents are sampled without replacement from the top 200 documents for that topic using a fixed random seed. Each document is first whitespace-normalized and then truncated to at most 1000 model tokens; if tokenization support is unavailable, the implementation falls back to whitespace-token truncation. We do not apply an additional character cap.

We use OpenAI \texttt{gpt-5-2025-08-07} through the Responses API. For LLM-generated descriptors, we use \texttt{reasoning\_effort=low}, \texttt{verbosity=low}, JSON-object output mode, and a maximum output budget of 3000 tokens. We leave temperature unspecified. If the response cannot be parsed into exactly 10 descriptors, we retry up to three times. The prompt asks the model to produce exactly 10 concise, non-redundant descriptors grounded only in the provided excerpts.

\begin{figure}[H]
\centering
\begin{tcolorbox}[
    colback=gray!5,
    colframe=gray!60,
    title=\textbf{Free-form LLM descriptor prompt},
    arc=2mm,
    boxrule=1pt
]
\small
You are given excerpts from documents that are highly associated with the same latent topic.\\
Task: Produce exactly 10 concise descriptors for this topic.\\
Rules: Each descriptor must be 4 to 10 words. Descriptors should be specific, non-redundant, and grounded only in the excerpts. Do not output a single topic name; output a ranked list of descriptors. Return JSON only, with this exact schema: \texttt{\{"descriptors": ["...", "..."]\}}.\\
Topic id: \texttt{<topic>}\\
Documents: \texttt{<document excerpts>}
\end{tcolorbox}
\end{figure}

\paragraph{MonoTM descriptors.}
For MonoTM, we use the topic--feature distribution $B$ estimated by the fixed-$\Theta$ descriptor model described in Appendix~\ref{sec:MonoTM-details}. We retain validated SAE features with $\mathrm{IS}(f)\ge 0.8$ and report the top 10 feature labels under $\beta_{t,f}$ for each topic. Unlike the free-form LLM descriptor baseline, MonoTM does not generate each topic's descriptor list freely from top documents. Instead, the descriptor list is selected from a fixed validated feature vocabulary grounded in document-embedding activations.

\paragraph{Topic naming for compact qualitative comparison.}
For compact comparison in Tables~\ref{tab:rq3-topic-names-20ng-app}, \ref{tab:rq3-topic-names-wos}, and \ref{tab:rq3-topic-names-reuters-app}, we generate high-level topic names from each descriptor list. The naming model receives only the descriptor list and the descriptor method; it does not receive the gold label or the candidate-label list. For word-based methods, the naming prompt receives the 30 ranked terms. For the free-form LLM descriptor baseline and MonoTM, it receives the 10 semantic descriptors.

We use the same OpenAI model, \texttt{gpt-5-2025-08-07}, through the Responses API with \texttt{reasoning\_effort=low}, \texttt{verbosity=low}, JSON-object output mode, and a maximum output budget of 300 tokens. We again leave temperature unspecified and retry up to three times if the JSON response cannot be parsed.

\begin{figure}[H]
\centering
\begin{tcolorbox}[
    colback=gray!5,
    colframe=gray!60,
    title=\textbf{Topic naming prompt},
    arc=2mm,
    boxrule=1pt
]
\small
I have a ranked list of \texttt{<top words/phrases or semantic descriptors>} derived from a single latent topic.\\
Descriptor method: \texttt{<method>}\\
Ranked descriptors: \texttt{<descriptor list>}\\
Task: Identify the high-level topic that these descriptors likely came from.\\
Rules: Output a concise topic name in 1 to 3 words. Do not include quotes, explanations, bullets, or extra text. Return JSON only with this exact schema: \texttt{\{"name": "..."\}}.
\end{tcolorbox}
\end{figure}

\section{High-Level Topic Labels Across SAE Configurations}
\label{gpt_topic_label}

To provide an intuitive qualitative check of MonoTM's output, we take each topic's top-10 interpretable feature labels (ranked by $\beta_{t,f}$) and ask \texttt{gpt-4.1-2025-04-14} (temperature $0$) to produce a 1--3 word topic name using the prompt in Appendix~\ref{sec:descriptor-baseline-details}. Table~\ref{tab:gpt-topics-stacked} shows these names for three SAE configurations ($(N{=}1,K{=}8)$, $(N{=}3,K{=}16)$, $(N{=}5,K{=}32)$). For an informative comparison, we align inferred topics to gold categories using the process described in Section~\ref{sec:rq2_results}. In many cases, the generated topic labels closely paraphrase the matched gold labels.

\input{gpt_topics_stacked_table.tex}

\section{Additional Descriptor Comparison Results}
\label{app:rq3-topic-name-tables}

Tables~\ref{tab:rq3-topic-names-20ng-app}–\ref{tab:rq3-topic-names-reuters-app} report additional topic-name tables for the controlled descriptor comparison in Section~\ref{sec:descriptor-comparison}. As in the main text, all methods describe the same inferred topics under fixed document--topic mixtures $\Theta$, and reference labels are shown only for orientation. 

\begin{table*}[t]
\centering
\scriptsize
\setlength{\tabcolsep}{3pt}
\renewcommand{\arraystretch}{1.08}
\begin{tabularx}{\textwidth}{lYYYY}
\toprule
\textbf{Reference} & \textbf{MonoTM} & \textbf{$\theta$-TF-IDF} & \textbf{c-TF-IDF} & \textbf{LLM-generated} \\
\midrule
alt.atheism & Religious debate & Objective morality & Objective morality & Meta-ethics \\
comp.graphics & Computer Graphics & Image format conversion & Image file formats & Computer Graphics \\
comp.os.ms-windows.misc & PC hardware troubleshooting & Hard drive setup & PC hard drives & DOS disk configuration \\
comp.sys.ibm.pc.hardware & Technical troubleshooting & Lead-acid batteries & Lead-acid batteries & Lead-acid battery storage \\
comp.sys.mac.hardware & PC hardware troubleshooting & Macintosh hardware & Macintosh hardware & Macintosh video hardware \\
comp.windows.x & X11 troubleshooting & X11 programming & X11 Motif & X11 programming \\
misc.forsale & Tech classifieds & Comics for sale & Marvel Comics & Classifieds \\
rec.autos & Automotive discussion & Cars & Cars & Performance cars \\
rec.motorcycles & Motorcycle Culture & Motorcycles & Motorcycles & Motorcycle riding techniques \\
rec.sport.baseball & Baseball Analytics & Baseball & Baseball & Baseball sabermetrics \\
rec.sport.hockey & Ice hockey & NHL Playoffs & NHL hockey & NHL Playoffs \\
sci.crypt & Clipper Chip Debate & Clipper Chip & Clipper Chip & Clipper Chip \\
sci.electronics & Computer hardware & Electronic circuits & Electronic circuits & Clickless audio switching \\
sci.med & Medical skepticism & Candida yeast syndrome & Candida overgrowth & Candida overgrowth \\
sci.space & Space commercialization & Lunar exploration & Spaceflight & Lunar habitation prize \\
soc.religion.christian & Theological debates & Christianity & Christian theology & Biblical hermeneutics \\
talk.politics.guns & Waco siege & Waco siege & Waco siege & Waco Siege \\
talk.politics.mideast & Ethno-national conflicts & Armenian Genocide & Armenian Genocide & Armenian genocide controversy \\
talk.politics.misc & American culture wars & Gun rights & Gun rights & Gay rights debate \\
talk.religion.misc & Usenet discussions & Email & Usenet newsgroups & Mailing list administration \\
\bottomrule
\end{tabularx}
\caption{High-level topic names generated from descriptor outputs on 20 Newsgroups. All methods describe the same inferred topics under fixed document--topic mixtures $\Theta$. Reference labels are shown only for orientation and are not provided to the topic-naming model.}
\label{tab:rq3-topic-names-20ng-app}
\end{table*}

\begin{table*}[t]
\centering
\small
\setlength{\tabcolsep}{3pt}
\renewcommand{\arraystretch}{1.08}
\begin{tabularx}{\textwidth}{lYYYY}
\toprule
\textbf{Reference} & \textbf{MonoTM} & \textbf{$\theta$-TF-IDF} & \textbf{c-TF-IDF} & \textbf{LLM-generated} \\
\midrule
biochemistry  & Biomedical research & Cancer cell signaling & Molecular oncology & Signal transduction \\
Civil  & Civil and Environmental Engineering & Rainwater harvesting & Rainwater Harvesting & Rainwater harvesting \\
CS  & Computer Science Research & Big Data Computing & Big Data & Cloud and Distributed Systems \\
ECE  & Control Systems Engineering & Power converter control & Power Converter Control & Digital Inverter Control \\
MAE  & Computational mechanics & Composite Materials & Composite materials & Mechanical behavior of materials \\
Medical  & Clinical epidemiology & Male hypogonadism & Male hypogonadism & Male hypogonadism \\
Psychology   & Psychology research & Parenting and child behavior & Child Development & Prosocial Behavior \\
\bottomrule
\end{tabularx}
\caption{High-level topic names generated from descriptor outputs on Web of Science. All methods describe the same inferred topics under fixed document--topic mixtures $\Theta$. Reference labels are shown only for orientation and are not provided to the topic-naming model.}
\label{tab:rq3-topic-names-wos}
\end{table*}

\begin{table*}[t]
\centering
\scriptsize
\setlength{\tabcolsep}{3pt}
\renewcommand{\arraystretch}{1.08}
\begin{tabularx}{\textwidth}{lYYYY}
\toprule
\textbf{Reference} & \textbf{MonoTM} & \textbf{$\theta$-TF-IDF} & \textbf{c-TF-IDF} & \textbf{LLM-generated} \\
\midrule
earn & Earnings reports & Earnings Reports & Earnings reports & Corporate Earnings \\
acq & Mergers and Acquisitions & Bank mergers & Bank Mergers & Bank Mergers \\
money-fx & Money market interventions & Money market operations & Money market & Open market operations \\
grain & Agricultural Trade & Export Enhancement Program & Grain export subsidies & Export Enhancement Program \\
crude & OPEC oil market & Oil production & OPEC oil production & Ecuador oil crisis \\
trade & US-Japan Trade Disputes & US-Japan trade dispute & US-Japan trade dispute & U.S.-Japan trade dispute \\
interest & Interest Rate Changes & Prime rate & Interest rates & Prime rate changes \\
ship & Persian Gulf conflict & Persian Gulf Tanker War & Gulf Tanker War & Operation Nimble Archer \\
wheat & Corporate financial distress & Earnings reports & Earnings reports & Nonrecurring items \\
corn & Mergers and Acquisitions & Retail mergers and acquisitions & Business news & Consumer retail M\&A \\
oilseed & Financial reporting & Corporate earnings & Corporate earnings & Corporate earnings \\
sugar & Agricultural trade & Sugar crop production & Crop production & Global sugar production \\
dlr & International monetary policy & Louvre Accord & Louvre Accord & G7 currency coordination \\
gnp & Macroeconomic Indicators & Economic Growth & Macroeconomic Indicators & OECD economic outlook \\
coffee & International Coffee Trade & International coffee trade & International Coffee Agreement & International Cocoa Agreement \\
veg-oil & Trade and M\&A & Mergers and acquisitions & Mergers and acquisitions & HBJ takeover battle \\
money-supply & Federal Reserve operations & Open market operations & Open market operations & Open market operations \\
gold & Mining Industry Finance & Gold mining & Gold mining & Gold mining \\
nat-gas & Airline Mergers & Airline mergers & Airline mergers & TWA USAir takeover \\
livestock & US Agricultural Policy & Livestock futures & Livestock futures & Meatpacking labor disputes \\
\bottomrule
\end{tabularx}
\caption{High-level topic names generated from descriptor outputs on the 20 most frequent Reuters labels. Because Reuters is multi-label and imbalanced, reference labels should be read as orientation labels rather than exhaustive topic descriptions.}
\label{tab:rq3-topic-names-reuters-app}
\end{table*}

\section{Representative Descriptor Lists}
\label{app:descriptor-examples}

Table~\ref{tab:rq3-descriptor-examples-app} provides representative descriptor lists before topic-name compression for the controlled descriptor comparison in Section~\ref{sec:descriptor-comparison}. All methods describe the same fixed document--topic mixtures $\Theta$. The examples illustrate the different roles of the descriptor mechanisms: word-based methods expose lexical anchors, free-form LLM-generated descriptors often summarize salient document clusters, and MonoTM descriptors are selected from a fixed vocabulary of validated SAE features ranked by a model-estimated topic--feature distribution.

\begin{table*}[t]
\centering
\scriptsize
\setlength{\tabcolsep}{4pt}
\renewcommand{\arraystretch}{1.12}
\begin{tabularx}{\textwidth}{llY}
\toprule
\textbf{Reference} & \textbf{Method} & \textbf{Top descriptors} \\
\midrule
\texttt{sci.crypt} & MonoTM &
Distrust of government and defense of liberty; government skepticism, privacy concerns, argumentative discourse; technical discourse on cryptographic algorithms and implementation; secure voice tech, compression, early internet security; government surveillance and cryptographic privacy concerns \\
\texttt{sci.crypt} & $\theta$-TF-IDF &
clipper; encryption; keys; escrow; chip; government; technology; secure; nsa; key \\
\texttt{sci.crypt} & c-TF-IDF &
encryption; clipper; chip; government; technology; key; keys; escrow; nsa; secure \\
\texttt{sci.crypt} & LLM-generated &
Government key-escrowed Clipper Chip proposal; secret Skipjack algorithm and family key; law enforcement access via escrowed unit keys; LEAF transmits session key encrypted for authorities; critics warn of mandated, tappable encryption standard \\
\midrule
\texttt{talk.politics.misc} & MonoTM &
Distrust of government and defense of liberty; contentious debate surrounding homosexuality and morality; moral debates about sexuality and statistics; gun control, self-defense arguments, statistical debate; contentious social debate, group identity, and conflict \\
\texttt{talk.politics.misc} & $\theta$-TF-IDF &
cramer; optilink; clayton; homosexuals; people; gay; gun; militia; rights; sexual \\
\texttt{talk.politics.misc} & c-TF-IDF &
people; cramer; edu; optilink; militia; homosexuals; gay; gun; sexual; rights \\
\texttt{talk.politics.misc} & LLM-generated &
Debating Kinsey versus Guttmacher homosexuality prevalence statistics; claims homosexuals exaggerate numbers to sway politicians; arguments over gay rights anti-discrimination employment laws; libertarian stance: private hiring without government interference; accusations linking gays with child molestation, NAMBLA \\
\midrule
\texttt{ECE} & MonoTM &
Engineered control systems and PID control design; electrical power system analysis and control; high-performance electric motor design and analysis; analog circuit design and performance metrics; engineered system optimization and energy efficiency analysis \\
\texttt{ECE} & $\theta$-TF-IDF &
converter; control; digital; voltage; proposed; power; current; controller; circuit; system \\
\texttt{ECE} & c-TF-IDF &
control; proposed; voltage; converter; power; current; controller; circuit; digital; system \\
\texttt{ECE} & LLM-generated &
FPGA-based digital control for power converters; digital PLLs for single-phase grid synchronization; discrete-time modulation introducing transport delays; finite control set MPC implemented fully on FPGA; deadbeat predictive current control in inverters \\
\midrule
\texttt{Civil} & MonoTM &
Applied hydrology, water resource engineering, rainwater harvesting; engineered system optimization and energy efficiency analysis; sustainable building and environmental responsibility concepts; construction project management, analytical methodology, performance evaluation; water pollution assessment and scientific reporting \\
\texttt{Civil} & $\theta$-TF-IDF &
water; rainwater; rwh; harvesting; demand; tank; system; saving; efficiency; supply \\
\texttt{Civil} & c-TF-IDF &
water; rainwater; rwh; harvesting; use; demand; tank; saving; efficiency; system \\
\texttt{Civil} & LLM-generated &
Rainwater harvesting supplying toilets, laundry, irrigation; economic feasibility via payback and NPV analyses; optimal storage tank sizing using daily simulations; water saving efficiency and reliability metrics; integrating rainwater with greywater reuse systems \\
\midrule
\texttt{crude} & MonoTM &
Oil price stability and OPEC market control; OPEC oil production, pricing, and market regulation; OPEC oil price and production control; OPEC oil production, pricing, and regulation reports; oil market data and energy economics \\
\texttt{crude} & $\theta$-TF-IDF &
bpd; oil; ecuador; opec; barrels; crude; prices; pipeline; production; exports \\
\texttt{crude} & c-TF-IDF &
said; oil; bpd; opec; ecuador; crude; barrels; pipeline; production; prices \\
\texttt{crude} & LLM-generated &
Earthquake halts Ecuador crude exports, declares force majeure; Lago Agrio--Balao pipeline severely damaged, costly repairs; temporary pipeline link to Colombia's Tumaco port; Venezuela lends Ecuador crude for exports, domestic needs; Ecuador plans over-quota OPEC production to repay loans \\
\midrule
\texttt{ship} & MonoTM &
Government economic intervention and trade policy; US-Iran military conflict and retaliation narratives; international military conflict and geopolitical tension; Persian Gulf naval conflict and escalation threat; Persian Gulf oil security, U.S. military response \\
\texttt{ship} & $\theta$-TF-IDF &
iran; gulf; iranian; attack; kuwaiti; oil; reflagged; ship; tanker; platform \\
\texttt{ship} & c-TF-IDF &
iran; said; gulf; iranian; oil; attack; kuwaiti; ship; platform; tanker \\
\texttt{ship} & LLM-generated &
U.S. destroyers shell Iranian Rostam oil platform; retaliation for Silkworm strike on Sea Isle City; Navy raids second platform, destroys radar communications; Iran vows crushing retaliation for platform attacks; escorting reflagged Kuwaiti tankers through Gulf \\
\bottomrule
\end{tabularx}
\caption{Longer representative descriptor lists before topic-naming. Word-based descriptors often expose terms or named entities; LLM-generated descriptors often describe salient document clusters; MonoTM descriptors are selected from validated SAE features and ranked by a topic--feature distribution.}
\label{tab:rq3-descriptor-examples-app}
\end{table*}

\section{Affordance Demonstration: Topic Relations in a Shared Semantic-Feature Space}
\label{app:semantic-feature-affordance}

This appendix demonstrates a downstream affordance of MonoTM's shared semantic-feature representation. The goal is not to introduce a new quantitative benchmark, but to illustrate how MonoTM can support analyses beyond compact topic description. Because MonoTM represents both documents and topics in a shared vocabulary of validated SAE features, users can inspect which features are shared across documents, which features distinguish neighboring topics, and which semantic attributes cut across topic boundaries. We demonstrate this affordance by analyzing how topics are related in the Web of Science corpus.

\paragraph{Construction details.}
We perform the analysis on Web of Science under the same fixed document--topic mixtures $\Theta$ used in the controlled descriptor comparison. For MonoTM, we use the estimated topic--feature distribution $B$ and retain the top 200 validated features per topic. We keep the raw $\beta_{t,f}$ masses rather than renormalizing them. For the lexical analogue, we use the top 50 terms per topic from the $\theta$-weighted TF-IDF topic--term representation, with term scores normalized to sum to one within each topic.

For two topics $i$ and $j$ with MonoTM topic--feature distributions $\beta_i$ and $\beta_j$, we define their shared feature mass as
\[
O(i,j)=\sum_f \min(\beta_{i,f},\beta_{j,f}).
\]
More generally, for either representation, we compute topic-pair overlap as
\[
O(i,j)=\sum_x \min(v_{i,x}, v_{j,x}),
\]
where $x$ indexes validated SAE features for MonoTM and lexical terms for the TF-IDF analogue. To explain each edge, we rank shared items by $\min(v_{i,x},v_{j,x})$ and report the top shared features or terms. The visualized networks display the eight strongest topic-pair overlaps, and the tables below report representative shared items from those edges.

\begin{figure*}[t]
  \centering
  \includegraphics[width=2\columnwidth]{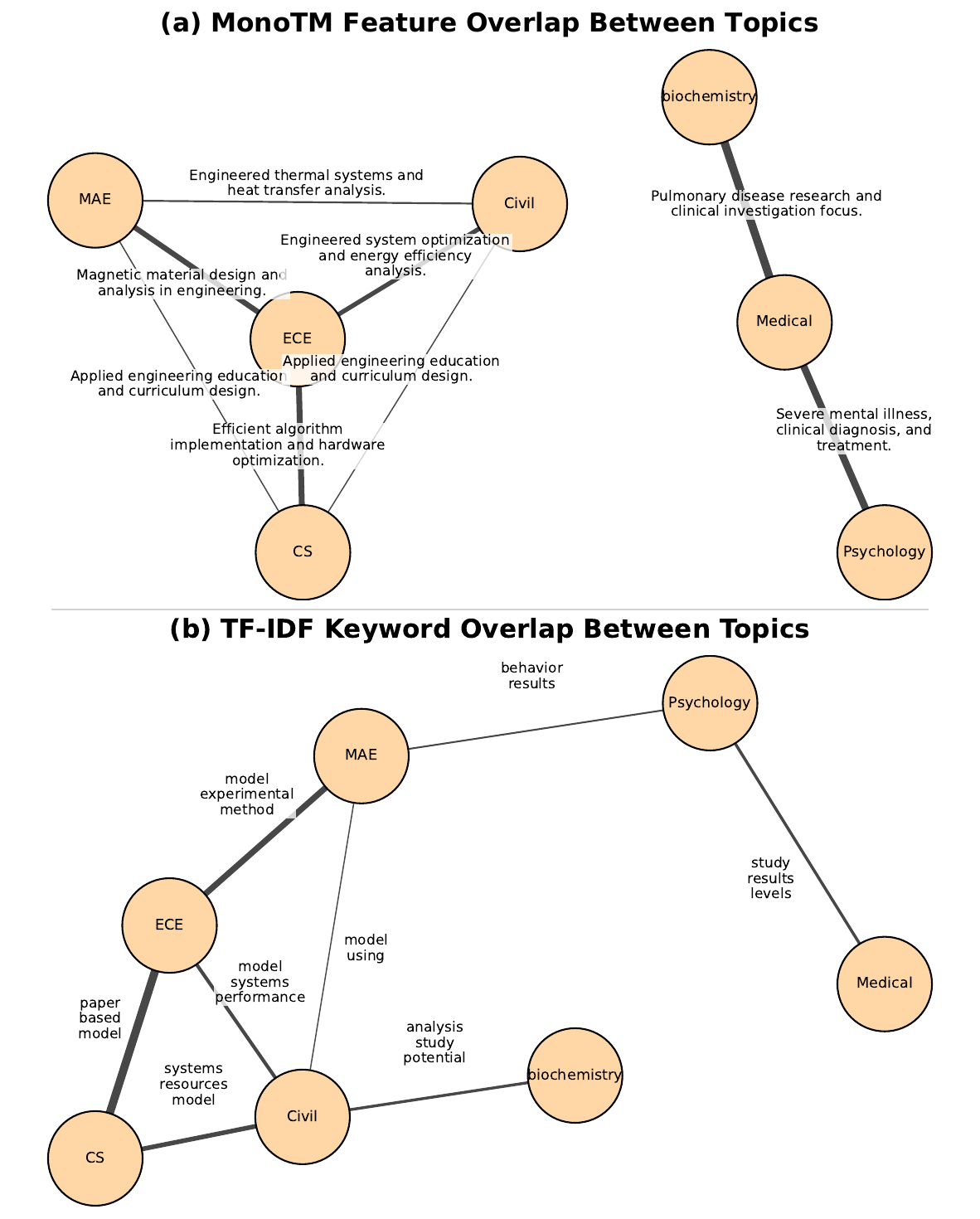}
  \caption{
  Web of Science topic relation networks under fixed document--topic mixtures.
  Edges show the strongest topic-pair overlaps, with edge width proportional to shared mass.
  (a) MonoTM computes overlap in the validated SAE-feature space, so edge labels identify the semantic features that explain each topic connection.
  (b) A $\theta$-weighted TF-IDF analogue computes overlap in a lexical term space, so edge labels identify shared keywords.
  The comparison illustrates how MonoTM supports feature-level interpretation of topic relations beyond lexical overlap.
  }
  \label{fig:topic-relation-networks}
\end{figure*}

\begin{figure*}[t]
  \centering
  \includegraphics[width=.48\textwidth]{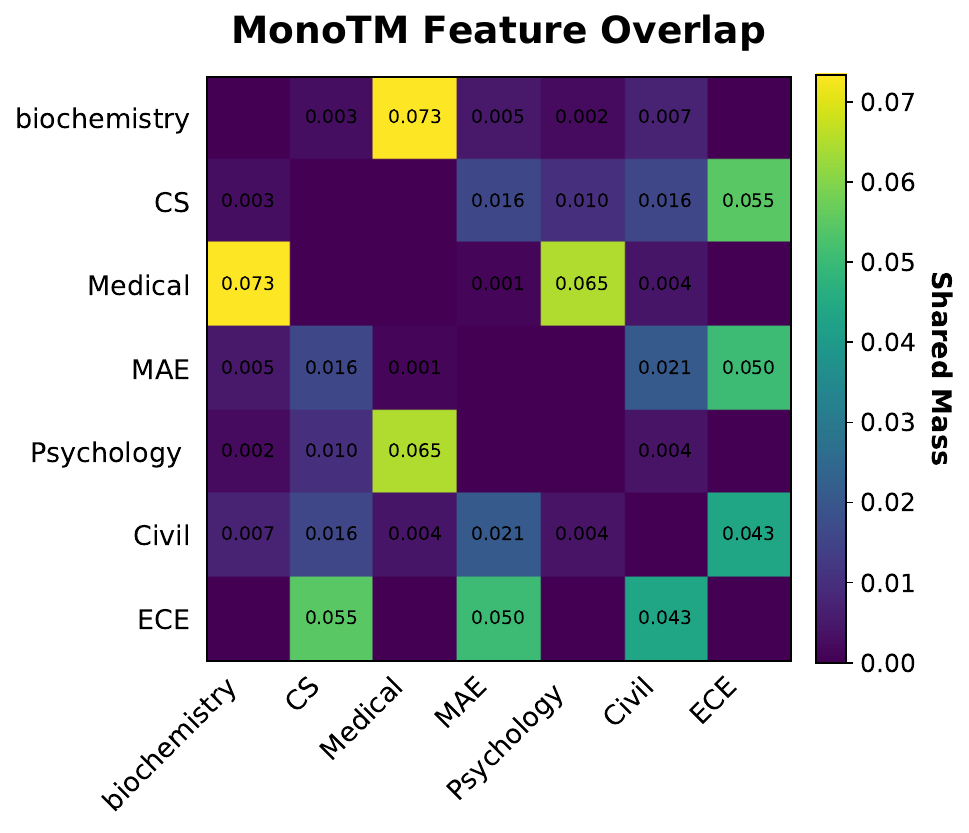}
  \hfill
  \includegraphics[width=.48\textwidth]{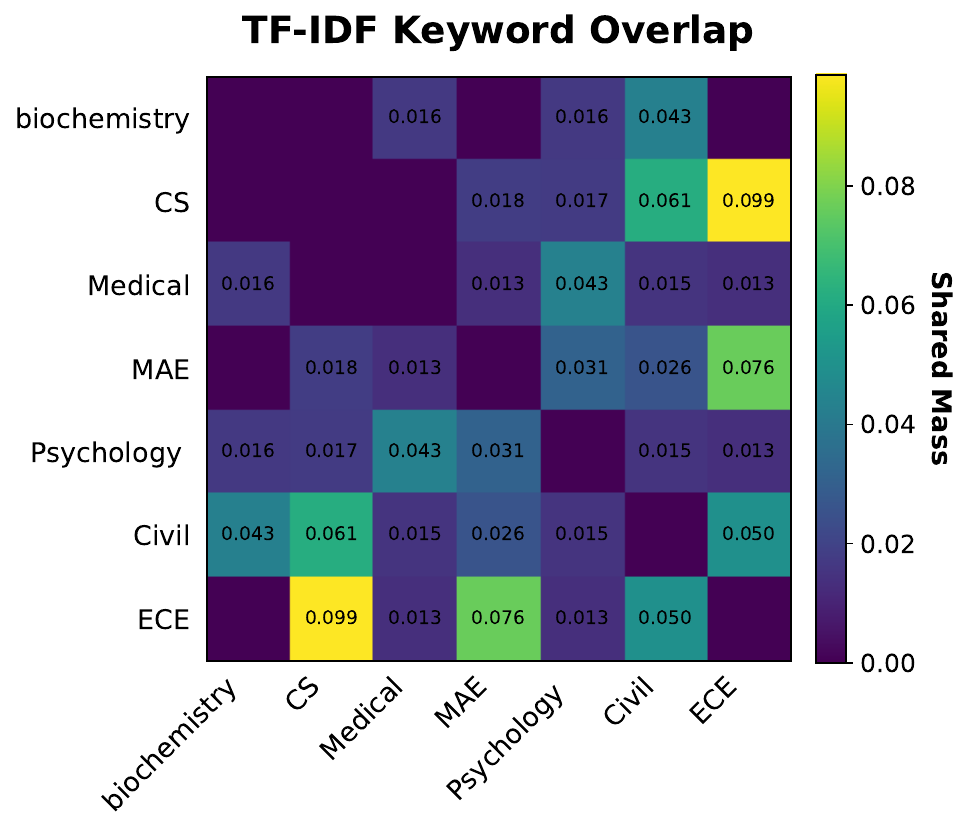}
  \caption{
  Topic-pair overlap matrices for Web of Science. Left: MonoTM shared-feature mass using the top 200 validated features per topic from $B$. Right: $\theta$-weighted TF-IDF shared-term mass using the top 50 lexical terms per topic. Diagonal entries are set to zero.
  }
  \label{fig:topic-relation-heatmaps}
\end{figure*}

To visualize the resulting topic relations, Figure~\ref{fig:topic-relation-networks} shows the strongest topic-pair overlaps as networks for MonoTM and the $\theta$-weighted TF-IDF analogue. Figure~\ref{fig:topic-relation-heatmaps} provides the corresponding full pairwise overlap matrices, showing whether the strongest network edges reflect isolated pairwise relations or broader overlap structure across topics.

\paragraph{Qualitative comparison.}
Figures~\ref{fig:topic-relation-networks} and~\ref{fig:topic-relation-heatmaps} show that the two representations yield qualitatively different topic-relation structures and edge explanations. Table~\ref{tab:topic-relation-shared-items} reports representative shared items for several high-overlap topic pairs. MonoTM links topics through semantic features such as clinical disease investigation, severe mental illness and treatment, efficient algorithm implementation and hardware optimization, applied engineering education, and engineered system optimization and energy efficiency. By contrast, the $\theta$-weighted TF-IDF analogue often explains topic links using lexical items such as \textit{model}, \textit{systems}, \textit{method}, \textit{study}, \textit{results}, \textit{paper}, and \textit{based}. Thus, a word-based representation can indicate that two topics overlap lexically, but MonoTM can make the semantic basis of the relation more explicit.

\begin{table*}[t]
\centering
\scriptsize
\setlength{\tabcolsep}{4pt}
\renewcommand{\arraystretch}{1.15}
\begin{tabularx}{\textwidth}{lYY}
\toprule
\textbf{Topic pair} & \textbf{MonoTM shared validated features} & \textbf{$\theta$-TF-IDF shared terms} \\
\midrule
biochemistry--Medical & Pulmonary disease research and clinical investigation focus.; Cancer research, molecular mechanisms, and clinical outcomes.; Gastrointestinal distress, IBS/IBD, patient well-being & study \\
Medical--Psychology & Severe mental illness, clinical diagnosis, and treatment.; Anxiety, depression, and psychological distress research.; Child development, family well-being, healthcare support. & study; results; levels \\
CS--ECE & Efficient algorithm implementation and hardware optimization.; Applied technology for infrastructure and efficiency.; Mathematical modeling and algorithmic development in data science & paper; based; model; performance; proposed \\
MAE--ECE & Magnetic material design and analysis in engineering.; Precise sensor measurement and instrumentation details.; Detailed thermodynamic system analysis and optimization. & model; experimental; method; using; results \\
Civil--ECE & Engineered system optimization and energy efficiency analysis.; Engineered thermal systems and heat transfer analysis.; Applied technology for infrastructure and efficiency. & model; systems; performance; using \\
MAE--Civil & Engineered thermal systems and heat transfer analysis.; Applied engineering education and curriculum design.; Technical design, optimization, and performance analysis. & model; using \\
\bottomrule
\end{tabularx}
\caption{Representative shared items explaining Web of Science topic-pair edges. MonoTM edge explanations are validated SAE features ranked by shared $\beta$ mass, while the lexical analogue explains edges through shared $\theta$-weighted TF-IDF terms.}
\label{tab:topic-relation-shared-items}
\end{table*}

Overall, this demonstration illustrates a broader affordance of MonoTM: the same validated feature vocabulary can support topic description, topic comparison, and inspection of semantic attributes that cut across benchmark category boundaries.

\end{document}

%% file: gpt_topics_stacked_table.tex

\begin{table*}[t] 
\centering
\small
\setlength{\tabcolsep}{4pt}
\renewcommand{\arraystretch}{1.15}
\newcolumntype{Y}{>{\raggedright\arraybackslash}X}
\begin{tabularx}{\textwidth}{lYYY}
\toprule
\textbf{Gold Label} & \textbf{N=1, K=8} & \textbf{N=3, K=16} & \textbf{N=5, K=32} \\
\midrule
\multicolumn{4}{l}{\textbf{20 Newsgroups}} \\
\midrule
rec.sport.hockey & Online Hockey Discussions & NHL Hockey Fans & Professional Hockey Discourse \\
soc.religion.christian & Christian Theology Debates & Religious Controversies & Religion and Morality \\
rec.motorcycles & Motorcycle Riding & Motorcycle Safety & Motorcycle Culture \\
rec.sport.baseball & Baseball Online Communities & Baseball Discussion & Baseball Analysis \\
sci.crypt & Cypherpunk Movement & Cryptography and Surveillance & Clipper Chip Controversy \\
\addlinespace[0.6em]
\midrule

\multicolumn{4}{l}{\textbf{Web of Science}} \\
\midrule
Medical  & Human Disease Research & Human Health Research & Human Health Disorders \\
Psychology   & Child and Adolescent Psychology & Psychological Research & Social Psychology \\
CS  & Computer Science & Computer Science & Computer Science \\
biochemistry  & Biomedical Science & Biomedical Research & Biomedical research \\
ECE  & Control Engineering & Electrical Engineering & Electrical Engineering \\
\addlinespace[0.6em]
\midrule
\multicolumn{4}{l}{\textbf{Reuters}} \\
\midrule
earn & Quarterly Financial Reporting & Financial Reporting & Financial Performance Comparison \\
acq & Mergers and Acquisitions & Financial News Events & Mergers and Acquisitions \\
money-fx & Central Bank Operations & Bank of England Interventions & Central Bank Interventions \\
grain & Agricultural Trade Markets & Agricultural Trade Reporting & Agricultural Trade \\
crude & Global Oil Markets & Global Oil Markets & OPEC Oil Markets \\
\bottomrule
\end{tabularx}
\caption{Topic labels generated from MonoTM outputs across three SAE configurations. The gold labels are shown only for orientation and are not provided during topic naming.}
\label{tab:gpt-topics-stacked}
\end{table*}